\documentclass[twoside]{article}

\usepackage[accepted]{aistats2026}
\usepackage[round]{natbib}

\usepackage{microtype}
\usepackage{amsmath}
\usepackage{amsfonts}
\usepackage{bm}

\usepackage{graphicx}
\usepackage{booktabs}
\usepackage{makecell}
\usepackage{nicefrac}

\usepackage{enumitem}
\usepackage{xcolor}
\usepackage{float}
\usepackage{wrapfig}
\usepackage[linesnumbered,ruled]{algorithm2e}
\usepackage{algpseudocode}
\usepackage{algorithmicx} 

\usepackage{float}      % Allows [H] placement specifier
\usepackage{caption}    % Customizing captions
\usepackage{subcaption}
\usepackage{mathabx}
\newcommand{\method}{H-learner}
\usepackage[colorlinks=true,linkcolor=blue,citecolor=blue,urlcolor=blue]{hyperref}
\begin{document}



\runningtitle{Hybrid Meta-learners for Estimating Heterogeneous Treatment Effects}

\twocolumn[

\aistatstitle{Hybrid Meta-learners for Estimating \\ Heterogeneous Treatment Effects}

% \vspace{-0.1in}
\aistatsauthor{ Zhongyuan Liang \And Lars van der Laan \And Ahmed Alaa }

\aistatsaddress{%
UC Berkeley, UCSF \And
University of Washington \And
UC Berkeley, UCSF
} 
]

\begin{abstract}
Estimating conditional average treatment effects (CATE) from observational data involves modeling decisions that differ from supervised learning, particularly concerning how to regularize model complexity. Previous approaches can be grouped into two primary ``meta-learner'' paradigms that impose distinct inductive biases. {\it Indirect} meta-learners first fit and regularize separate potential outcome (PO) models and then estimate CATE by taking their difference, whereas {\it direct} meta-learners construct and directly regularize estimators for the CATE function itself. Neither approach consistently outperforms the other across all scenarios: indirect learners perform well when the PO functions are simple, while direct learners outperform when the CATE is simpler than individual PO functions. In this paper, we introduce the {\it Hybrid Learner} (H-learner), a novel regularization strategy that interpolates between the direct and indirect regularizations depending on the dataset at hand. The H-learner achieves this by learning intermediate functions whose difference closely approximates the CATE without necessarily requiring accurate individual approximations of the POs themselves. We demonstrate that intentionally allowing suboptimal fits to the POs improves the bias-variance tradeoff in estimating CATE. Experiments conducted on semi-synthetic and real-world benchmark datasets illustrate that the H-learner consistently operates at the Pareto frontier, effectively combining the strengths of both direct and indirect meta-learners.
\noindent\textbf{Code:} \url{https://github.com/AlaaLab/H-learner}
\end{abstract}

\section{Introduction}
\label{sec:intro}

% \vspace{-.05in}
We consider the problem of estimating conditional average treatment effects (CATE) from observational data \citep{johansson2016learning,alaa2017bayesian, shalit2017estimating,wager2018estimation,kunzel2019metalearners,kennedy2020towards}. Let $\mathcal{D} = \{(X_i, T_i, Y_i)\}_{i=1}^n$ denote the observational dataset, where $X_i \in \mathcal{X}$ is a feature vector, $T_i \in \{0, 1\}$ is a binary treatment indicator and $Y_i \in \mathbb{R}$ is the observed outcome of interest. Under the Neyman-Rubin potential outcomes framework \citep{neyman1923application}, let $Y_i(1)$ and $Y_i(0)$ denote the potential outcomes (POs) for individual $i$ under treatment and control, respectively.~The {\it fundamental problem of causal inference} is that we only observe one of the POs~for~each~individual, i.e., $Y_i = T_i Y_i(1) + (1 - T_i) Y_i(0)$. We denote $\mu_0(x) = \mathbb{E}[Y(0) \mid X=x]$ and $\mu_1(x) = \mathbb{E}[Y(1) \mid X=x]$ as the expected POs under control and treatment. Our goal is to estimate the CATE function:
% \vspace{-0.08in}
\begin{equation}
\tau(x) = \mathbb{E}[Y(1) - Y(0) \mid X = x],
\end{equation}
using the dataset $\mathcal{D}$. This estimand is identifiable from the sample $\mathcal{D}$ under the following assumptions: {\it (1) Consistency:} $Y_i = Y_i(t)$, {\it (2) Unconfoundedness:} $(Y(0), Y(1)) \perp\!\!\!\perp T \mid X$, and {\it (3) Positivity:} $ \forall x \in \mathcal{X}, \  0 < \pi(x) < 1$, where $\pi(x) = \mathbb{P}(T = 1 \mid X = x)$ is the propensity score \citep{rubin1974estimating}.

\begin{figure*}[t]
  \centering
  \includegraphics[width=\textwidth]{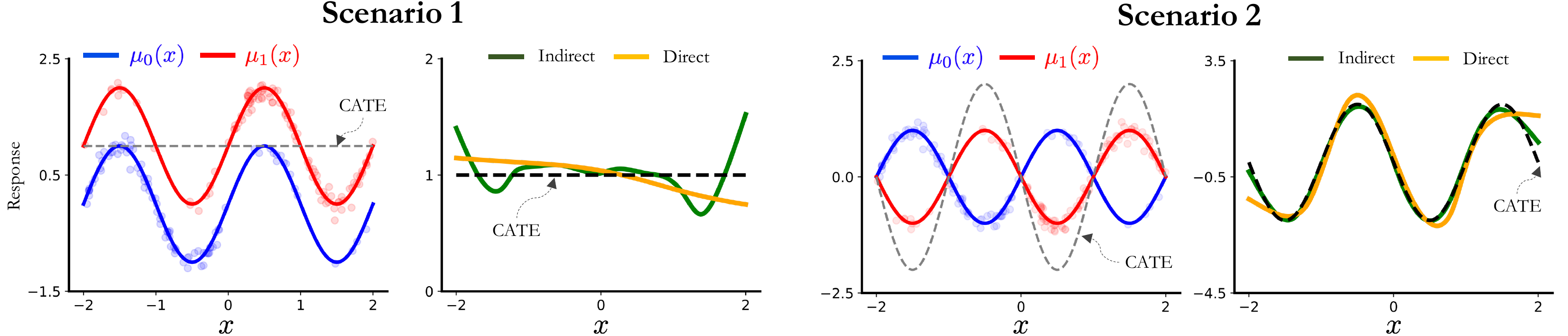}
  \caption{\textbf{Impact of inductive biases on CATE estimation.} (a) When CATE is simpler than POs, indirect learners introduce spurious heterogeneity due to independent regularization. (b) When CATE is as complex as POs, indirect learners perform better by accurately modeling each outcome surface.}
  \label{fig:1}
  \vspace{-.035in}
  \rule{\textwidth}{0.5pt}
  % \vspace{-.325in}
\end{figure*}

Learning the CATE function differs from standard supervised learning in several important ways. As noted by \citet{curth2021inductive}, two key distinctions stand out: (1) {\it covariate shift} induced by confounding, where the feature distributions differ across treatment and control groups, i.e., $\mathbb{P}(X \mid T=1) \neq \mathbb{P}(X \mid T=0)$, and (2) the {\it inductive biases} associated with the goal of estimating a difference between two outcome functions, $\tau(x) = \mu_1(x) - \mu_0(x)$, rather than the individual POs, which introduces unique considerations in how we design and regularize the models for $\mu_0(x)$, $\mu_1(x)$, and $\tau(x)$. In this paper, we focus on the second aspect and propose a new strategy for regularizing CATE models.

To understand how regularization affects learning of CATE, consider the following illustrative example, with variants of this example appear in the literature \citep{kunzel2019metalearners, kennedy2020towards, curth2021inductive}. We define the POs as:
\begin{align}\label{toymodel}
\mu_0(x) = \sin(\omega x), \quad \mu_1(x) = \sin(\omega x + \Delta) + \beta.
\end{align}
Here, the control and treatment outcome functions share the same sinusoidal shape and frequency, and differ only by a phase shift $\Delta$ and an amplitude shift $\beta$. As a result, both $\mu_0(x)$ and $\mu_1(x)$ have comparable complexity, i.e., both are equally easy—or difficult—to learn. However, the complexity of the resulting treatment effect function $\tau(x) = \mu_1(x) - \mu_0(x)$ can vary significantly depending on the values of $\Delta$. We highlight this by considering the following two scenarios:

$\bullet$\, {\bf Scenario 1:} If $\Delta = 0$, the phase alignment implies the treatment effect is constant across all units: $\tau(x) = \mu_1(x)-\mu_0(x) = \beta, \forall x \in \mathbb{R}$. In this case, $\tau(x)$ is much simpler than either PO, both~of~which are nonlinear. This scenario is common in medicine, where prognostic factors (predicting outcomes) may be complex, while predictive factors (modulating treatment effects) are simpler or fewer.

$\bullet$\, {\bf Scenario 2:} If $\Delta > 0$, the treatment effect becomes heterogeneous: $\tau(x) = \mu_1(x)-\mu_0(x) = \beta + (\sin(\omega x + \Delta) - \sin(\omega x))$. Now, $\tau(x)$ inherits the full complexity of the POs.~Though~the~individual POs have not changed in complexity, the induced CATE function is substantially more complex.

This example illustrates that the intrinsic difficulty in learning the treatment effect depends not just on the complexity of the POs, but also on the relationship between them. As such, choosing how to regularize CATE estimators requires careful consideration of which function—outcome or effect—can be better estimated in the given domain. 

\citet{kunzel2019metalearners} coined the notion of ``meta-learners'' to describe a family of model-agnostic strategies for learning CATE. Broadly, these strategies fall into two~categories~\citep{curth2021inductive}:

{\bf Indirect meta-learners} use the datasets $\mathcal{D}_0 = \{(X_i, Y_i): T_i=0\}$ and $\mathcal{D}_1 = \{(X_i, Y_i): T_i=1\}$ to obtain intermediate models of the PO functions $\widehat{\mu}_t, t \in \{0, 1\},$ and then set $\widehat{\tau}(x) = \widehat{\mu}_1(x) - \widehat{\mu}_0(x)$. 

{\bf Direct meta-learners} target the CATE function directly by constructing a {\it pseudo-outcome}~$Y_\varphi$ based~on some nuisance parameter $\varphi$ (e.g., estimates of the propensity score $\widehat\pi(x)$ and the POs $\widehat{\mu}_t$, $t \in \{0, 1\}$), and then fit a model $\widehat{\tau}(x)$ for the CATE function using the constructed dataset $\{(X_i, Y_{\varphi, i} )\}$. 

Figure \ref{fig:1} shows how direct and indirect learners perform under both scenarios. In Scenario~1, where the CATE function is simpler than the PO, the direct learner outperforms the indirect approach. This is~because indirect learners separately fit and regularize the PO models, and each model make errors in different regions of the covariate space. When the two models are subtracted to estimate the treatment effect, they can compound in unpredictable ways, introducing spurious heterogeneity into the CATE function. This phenomenon is known as ``regularization-induced confounding'' \citep{hahn2018regularization}. Conversely, in Scenario 2, where the CATE function is as complex as the POs, direct learners tend to underperform, as the benefit of targeting a simpler effect function no longer holds. In practice, however, it is difficult to know in advance which strategy will perform better, as their relative performance also depends on other factors underlying data generation processes (DGPs), including sample sizes, treatment imbalance, the degree of confounding and overlap.

To address these limitations, we propose the {\it Hybrid learner} (H-learner), a novel meta-learner regularization strategy for CATE estimation that generalizes both direct and indirect meta-learners. Our strategy formulates a proximal indirect learning task, where the model learns two intermediate functions, $f_0$ and $f_1$, which are not necessarily optimal estimators of the POs but are trained such that their difference accurately approximates the treatment effect. Regularization is applied by weighting two competing objectives: encouraging $f_0$ and $f_1$ to closely approximate the true POs, and ensuring that their difference $f_1 - f_0$ aligns with a pseudo-outcome-based guess of CATE.

% \begin{figure}[H]
% \centering
% \includegraphics[width=2.6in]{Figures/H-learner-Fig-2.pdf} 
% %\vspace{-.1in}
% \caption{Depiction of the PO and CATE~estimates based on the H-learner regularization strategy.}
% \label{fig:2}
% \end{figure}

% Following the taxonomy established in the literature, we refer to our method as the {\it Hybrid learner} (H-learner), as it combines elements of both direct and indirect regularization strategies. 

% Figure~\ref{fig:2} illustrates the performance of the H-learner under the two scenarios described earlier. In~Scenario~1, the H-learner identifies intermediate functions $\widehat f_0$ and $\widehat f_1$ that are not perfect fits to the POs,~but~whose difference is regularized by what a direct estimator would have produced. This leads~to~a~more~accurate CATE estimate than those obtained by either direct or indirect learners alone. In~Scenario~2,~the H-learner adapts to the more complex~treatment~effect structure by placing greater emphasis on accurately fitting the two POs, and continues to perform competitively, even when compared to an indirect learner, which focuses solely on outcome prediction.

In the remainder of the paper, we describe the proposed H-learner regularization strategy and place it in the context of existing meta-learning paradigms. Our method is model-agnostic and compatible with a wide range of machine learning (ML) architectures. In particular, we make the following contributions:

\vspace{-0.1in}

\begin{enumerate}[left=0pt, itemsep=0pt]

% \item We disentangle the inconsistent performance of meta-learners by identifying key factors of the underlying DGP, including functional complexity, confounding strength, treatment imbalance, and explicitly link these factors to the inductive bias introduced by the regularization strategies of meta-learners.

\item We characterize the performance of meta-learners by identifying key factors of the underlying DGPs and explicitly linking them to the inductive biases introduced by their regularization strategies.

\item We introduce the H-learner as a unified approach that combines direct and indirect regularization and theoretically demonstrate that it achieves lower prediction risk by balancing the bias–variance trade-off inherent in existing meta-learners.

% \item We introduce the H-learner, a principled meta-learner that unifies direct and indirect regularization and theoretically demonstrates that it overcomes the bias–variance trade-off inherent in existing learners.

\item We empirically show that the H-learner consistently achieves state-of-the-art performance, remains robust across diverse DGP characteristics, and lies on the Pareto frontier of common benchmarks.

\end{enumerate}

% In the remainder of the paper, we describe the proposed H-learner regularization strategy and place it in the context of existing meta-learning approaches for CATE estimation. Our method is model-agnostic and compatible with a wide range of machine learning architectures. We demonstrate that H-learners improve the bias–variance tradeoff in estimating CATE. Experiments on semi-synthetic and real-world datasets show that it consistently lies on the Pareto frontier by adapting the strengths of direct and indirect learners to different DGPs.
\section{Meta-Learners for Estimating Heterogeneous Treatment Effects}
\label{sec:metalearners}
% \vspace{-.05in}
In this section, we illustrate representative examples of indirect and direct meta-learners from the literature, and discuss the implicit inductive biases and regularization they impose.
\subsection{Indirect Meta-learners}
\label{sec:indirect}
% \vspace{-.05in}
Given an observational dataset partitioned by the two treatment groups, $\mathcal{D}_0$ and $\mathcal{D}_1$, an indirect meta-learner estimates the PO models $\mu_0$ and~$\mu_1$~by~independently fitting outcome models on each group by  minimizing a regularized empirical risk of the form:
% \vspace{-0.04in}
\begin{align}\label{indirect:loss}
\mbox{$\sum_{i=1}^n$} \ell\big(Y_i, \mu_{T_i}(X_i; \theta_{T_i})\big) + \lambda \, \mathcal{R}(\theta_0, \theta_1),
\end{align}
where $\ell$ is a supervised loss function (e.g., squared error), $\theta_0$ and $\theta_1$ are the parameters of the models for the control and treatment groups respectively, and $\mathcal{R}(\cdot)$ is a regularization term (e.g., $\ell_2$ norm, shared representation penalty, or complexity constraint). The final CATE estimate is then given by the difference between these plug-in estimates of the PO functions: $\widehat{\tau}(x) = \mu_1(x; \widehat{\theta}_{1}) - \mu_0(x; \widehat{\theta}_{0})$.

Several well-known methods fall under the category of indirect meta-learners. The {\it T-learner}~fits~two separate models $\mu_{0}(x; \theta_{0})$ and $\mu_{1}(x; \theta_{1})$ independently using the datasets $\mathcal{D}_0$ and $\mathcal{D}_1$, and estimates CATE as their difference. Another architecture is the {\it S-learner}, which fits a single model $\mu(x, t; \theta)$ by treating the treatment indicator as an additional feature, then estimating CATE by toggling this feature $t$, i.e., $\mu(x, 1; \widehat\theta)-\mu(x, 0; \widehat\theta)$. Examples of these models include Bayesian additive regression trees (BARTs), regression trees and representation-based methods such as TARNet, which learn a shared latent representation of the features before fitting separate heads for each treatment arm \citep{hill2011bayesian, athey2016recursive, shalit2017estimating}. 

However, because indirect learners optimize and regularize $\mu_0$ and $\mu_1$ independently, their inductive bias favors simpler functions that best fit $\mathcal{D}_0$ and $\mathcal{D}_1$. As a result, taking their difference can yield biased estimates and implicitly induce spurious treatment effect heterogeneity, particularly under DGPs where the CATE function is simpler than the POs \citep{hahn2018regularization, hahn2020bayesian}. This occurs because the models are not regularized with respect to the difference that defines the CATE, i.e., $\tau(x) = \mu_1(x) - \mu_0(x)$. \citet{curth2021inductive} further points out that this inductive bias not only risks introducing artefactual heterogeneity but also contradicts how treatment effect heterogeneity should be approached from a scientific standpoint—where the default (null) hypothesis is often that treatment effects are homogeneous \citep{crump2008nonparametric, ballman2015biomarker}.

\subsection{Direct Meta-learners}
\label{sec:direct}

Direct meta-learners are typically implemented as two-stage procedures. In the first stage, a nuisance parameter $\varphi$—such as the propensity score or outcome regression—is estimated and used to construct a pseudo-outcome $Y_\varphi$. In the second stage, a single model is trained to estimate ${\tau}(x)$ using the pseudo-labeled dataset $\{(X_i, Y_{\varphi,i})\}$. The pseudo-outcome is designed so that it~depends only on observables and satisfies, either exactly or approximately, the following condition:
\begin{align}\label{direct:learner}
E[Y_{\varphi} | X = x] = E[Y(1) - Y(0) | X = x],\, \forall x \in \mathcal{X}.
\end{align}
Once the pseudo-outcomes are constructed, the CATE is estimated via empirical risk minimization:
\begin{align}\label{direct: loss}
\mbox{$\sum_{i=1}^n$} \ell\big(Y_{\varphi,i}, \tau(X_i; \theta)\big) + \lambda \, \mathcal{R}(\theta),
\end{align}
where $\ell$ is a loss function and $\mathcal{R}(\theta)$ is a regularization applied directly to the CATE function. The general framework described above captures various methods in literature, including the X-learner, the Inverse propensity weighted (IPW) learner and the Doubly-robust (DR) learner (Table \ref{table:1}). When the propensity score is known, the IPW and DR pseudo-outcomes satisfy (\ref{direct:learner}) exactly. The regularization strategy underlying direct meta-learners differs fundamentally from that of indirect learners: rather than independently regularizing the PO models, direct learners apply training and regularization directly to the CATE function. 

However, the effectiveness of this approach hinges on the quality of the pseudo-outcome transformation, i.e., how accurately it reflects the CATE function, and it is largely determined by factors underlying the DGP. When the CATE surface is higher-dimensional than the underlying POs, direct learners often yield poorer estimates as they target a harder function. Imbalanced propensities can further worsen pseudo-outcome constructions—such as those based on inverse propensity—which suffer from high variance and degrade performance \citep{curth2021nonparametric}.

\vspace{-.05in}

\begin{table}[h]
\centering
\resizebox{\columnwidth}{!}{%
\begin{tabular}{rllll}
\specialrule{.2em}{.1em}{.1em}  \\[-4ex] 
& \\ 
\multicolumn{1}{l}{} & \multicolumn{4}{c}{\hspace{-9em}{\bf Pseudo-outcome}} \\ [0.5ex] \hline 
 &   \\[-1.5ex]
{\it IPW-learner} \citep{horvitz1952generalization} & \multicolumn{4}{l}{\hspace{-1em} $Y_{\varphi} =  \frac{T-\pi(X)}{\pi(X)(1-\pi(X))} Y$} \\[5pt] 
{\it X-learner} \citep{kunzel2019metalearners} & \multicolumn{4}{l}{\hspace{-1em} $Y_{\varphi} = T(Y-\widehat{\mu}_0(X)) + (1-T)(\widehat{\mu}_1(X)-Y)$} \\[5pt] 
{\it DR-learner} \citep{kennedy2020towards} & \multicolumn{4}{l}{\hspace{-1em} $Y_{\varphi} =  \frac{T-\pi(X)}{\pi(X)(1-\pi(X))} (Y-\widehat{\mu}_T(X)) + \widehat{\mu}_1(X) - \widehat{\mu}_0(X)$} \\ [5pt] 
\specialrule{.2em}{.1em}{.1em} 
\end{tabular}}
\caption{Direct meta-learners based on pseudo-outcome regression.}
% \vspace{-.2in}
\label{table:1}
\end{table}
\section{Hybrid Meta-learners}
\label{sec:hlearner}
\begin{figure}[!t]
    \centering
    \includegraphics[width=3.3in]{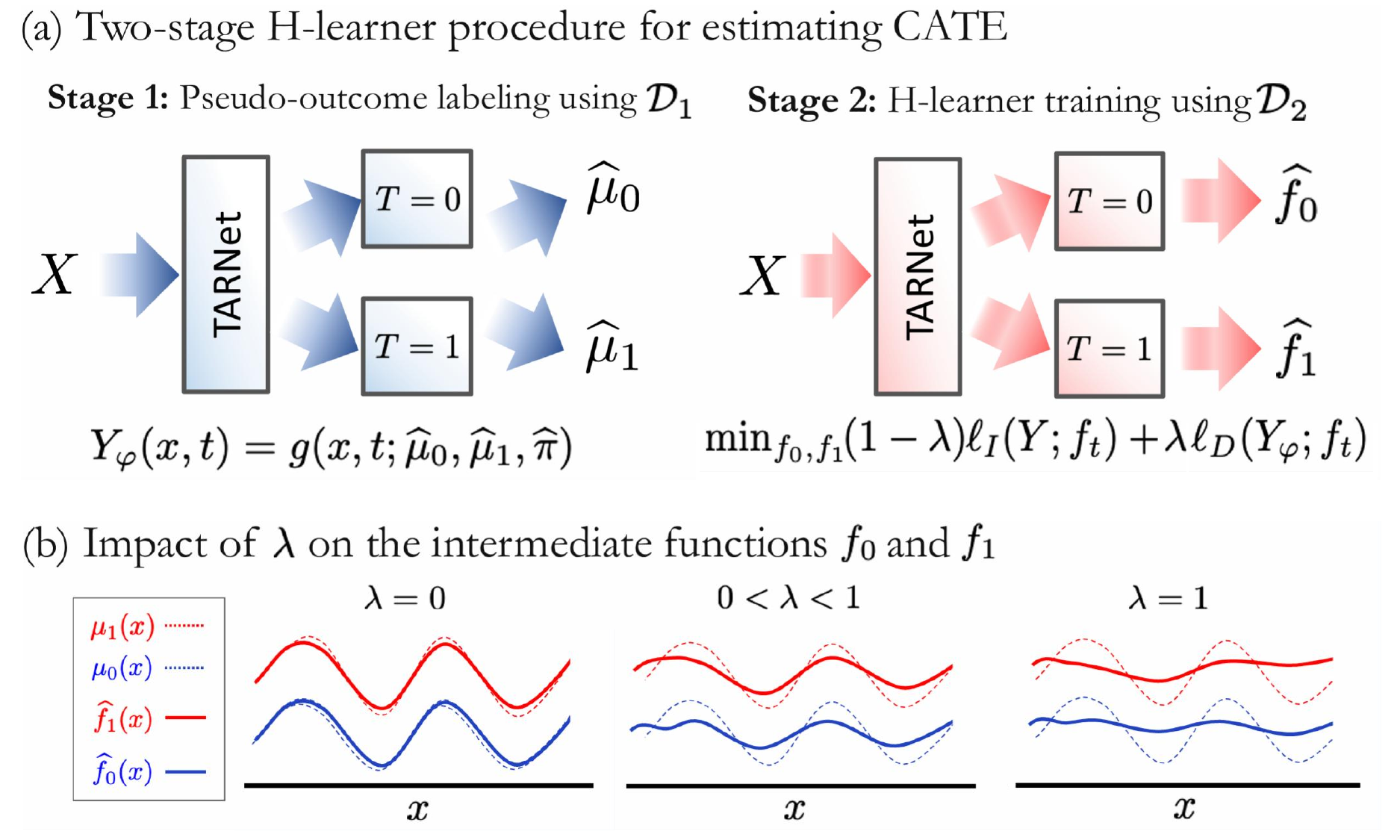}
    \caption{Pictorial depiction of the H-learner.}
    \label{fig:3}
\end{figure}

% \vspace{-.05in}
The key idea behind the proposed H-learner is to train a pair of models $f_0, f_1 \in \mathcal{F}$, where the CATE is estimated as their difference, $f_1(x) - f_0(x)$. However, unlike traditional indirect learners, which train $f_0$ and $f_1$ solely to minimize empirical risk with respect to the POs, the H-learner jointly optimizes these models to balance both accurate prediction of the POs and the accuracy of their implied CATE. As a result, the optimal finite-sample solutions $\widehat f_0, \widehat f_1 \in \mathcal{F}$ may not correspond to the best fits for the POs individually, but may instead favor a pair whose difference yields a more accurate estimate of the implied treatment effect function. Similar to direct meta-learners, the H-learner follows a two-stage procedure outlined below (Figure~\ref{fig:3}):

% The key idea behind the proposed H-learner is to train a pair of models $f_0, f_1 \in \mathcal{F}$, where the CATE is estimated as their difference, $f_1(x) - f_0(x)$, in a manner similar to indirect meta-learners. However, unlike traditional indirect learners, which train $f_0$ and $f_1$ solely to minimize empirical risk with respect to the POs, the H-learner jointly optimizes these models to balance both accurate prediction of the POs and the accuracy of their implied CATE. As a result, the optimal finite-sample solutions $\widehat f_0, \widehat f_1 \in \mathcal{F}$ may not correspond to the best fits for the POs individually, but may instead favor a pair whose difference yields a more accurate estimate of the implied treatment effect function. Similar to direct meta-learners, the H-learner follows a two-stage procedure outlined below:

{\bf Stage 1:} Construct a pseudo-outcome using a subset of the observational dataset $\mathcal{D}$ based on any of the transformations in Table \ref{table:1}. For instance, the X-learner pseudo-outcome can be constructed as:
\begin{align}\label{stage:1}
Y_{\varphi,i} = T_i (Y_i - \hat{\mu}_0(X_i)) + (1 - T_i)(\hat{\mu}_1(X_i) - Y_i),
\end{align}
where $\hat{\mu}_0$ and $\hat{\mu}_1$ are initial estimates of the POs for the control and treatment groups, respectively. These initial estimates can be obtained through any indirect learner, e.g., the T-learner described earlier. 

{\bf Stage 2:} Fit two intermediate functions \( f_0, f_1 \in \mathcal{F}\) by jointly minimizing the empirical risk~of~each function in predicting the observed (factual) outcomes, along with the empirical risk of their difference in approximating the pseudo-outcome constructed in {\bf Stage 1}. Formally, the H-learner loss function balances both POs prediction and treatment effect estimation, and is given by:

\begin{equation}\label{hlearner:loss}
\begin{aligned}
\min_{f_0, f_1}\, \mbox{$\sum_{i=1}^n$} \; &
(1-\lambda) \underbrace{\ell\!\left(Y_i, f_{T_i}(X_i)\right)}_{\text{Indirect component $\ell_I$}} \\[1pt]
&+ \lambda\,
\underbrace{\big((f_1(X_i)-f_0(X_i)) - Y_{\varphi,i}\big)^2}_{\text{Direct component $\ell_D$}},
\end{aligned}
\end{equation}

where $0 \leq \lambda \leq 1$, and we obtain the final CATE estimator as $\hat{\tau}(x) = \hat{f_1}(x) - \hat{f_0}(x)$. 

The H-learner loss can be viewed as a regularized version of the indirect learning framework, with a regularization term that encourages the difference between the two intermediate models to align with a pseudo-outcome. Alternatively, it can be interpreted as a generalized regularization strategy that interpolates between direct learners ($\lambda = 1$) and indirect learners ($\lambda = 0$) while adaptively balancing the influence of each. In the following sections, we demonstrate that the optimal value of $\lambda$ often depends on several factors of the underlying DGP, but in most cases, lies strictly within $\lambda \in (0,1)$ rather than at the boundary values. This reflects the benefit of leveraging both regularization strategies rather than relying on either one alone. Practically, we treat $\lambda$ as a hyperparameter and adaptively tune its value by minimizing the validation loss. Details of this validation procedure are provided in Appendix~\ref{sec:implementations}.

The H-learner procedure is compatible with any existing CATE estimation architecture that provides explicit estimates of the POs. In our implementation, we use the TARNet architecture both to estimate the nuisance components required for constructing pseudo-outcomes and to fit the intermediate models used to derive the final CATE estimate as illustrated in Figure~\ref{fig:3}(a). Further implementation details with a pseudo-algorithm can be found in the Appendix \ref{sec:implementations}.

\section{Related Work}
\label{sec:related}
% \vspace{-.05in}
\subsection{Inconsistent Performance of Indirect and Direct Learners}
\label{subsec:related1}
% \vspace{-.05in}
The divergence in performance between indirect and direct learners has been consistently observed across settings in the literature. In particular, prior work has shown that no single meta-learner consistently outperforms others across all settings \citep{kunzel2019metalearners}. Both theoretical and empirical studies show that their relative performance varies substantially with the sparsity and smoothness of the response surfaces \citep{curth2021nonparametric}. \citet{acharki2023comparison} extended the comparison of meta-learners to the non-binary treatment setting, theoretically highlighting the bias–variance trade-off in the multi-treatment setting. These inconsistencies highlight the importance of developing a more robust approach, as the true DGP is often unknown in practice, and relying on either method can result in unreliable performance.

% As discussed in Section~\ref{sec:metalearners}, the regularization strategies employed by direct and indirect learners reflect different inductive biases, leading to distinct strengths and weaknesses. 
% This divergence in performance has been consistently reported across the literature. In particular, prior work has shown that no single meta-learner consistently outperforms others across all settings \citep{kunzel2019metalearners}. Both theoretical and empirical studies demonstrate that their relative performance can vary significantly depending on the underlying DGPs, quantified by factors such as sparsity, smoothness, and the relative complexity of the POs and CATE functions \citep{curth2021nonparametric, curth2021inductive}. \citet{acharki2023comparison} further compared meta-learners in the multi-treatment setting, theoretically revealing a bias–variance trade-off shaped by their estimation strategies. These performance trade-offs highlight the importance of developing a more robust approach, as the true DGP is often unknown in practice, and relying on either method can result in unreliable performance.

\subsection{Regularization Strategies for CATE Estimations}
\label{subsec:related2}
% \vspace{-.05in}
Several other regularization strategies have been proposed to mitigate ``regularization-induced confounding''. \textit{Weight regularization} penalizes discrepancies between the weights of the~outcome heads to encourage parameter sharing \citep{hahn2018regularization}. \textit{Reparameterization} implicitly introduces regularization by modeling \( \mu_1(x) = \mu_0(x) + \tau(x) \), allowing the treatment effect  \( \tau(x) \) to be predicted directly by one head as an offset from \( \mu_0(x) \) \citep{imai2013estimating, curth2021inductive}. \textit{Structural regularization} was proposed with FlexTENet, a specialized neural architecture that enforces regularization through flexible feature sharing across layers \citep{curth2021inductive}. 
% These regularization strategies assume that reparameterization or feature sharing improves the performance by simplifying the learning problem.
% However, these approaches assume that reparameterization or feature sharing simplifies the learning problem, which holds only when the underlying CATE function is easier to estimate than the individual potential outcomes. Since the data-generating process is unknown in practice, naively applying these approaches can degrade performance when the assumption is violated.

\textbf{Comparison with H-learner:} \method\ formulation easily reveals the limitations of these regularization strategies. Penalizing weight differences between \( f_0 \) and \( f_1 \), or enforcing shared layers when implementing these functions with neural networks, effectively reduces the output difference \( f_1(x) - f_0(x) \), which is equivalent to applying \method\ with \( Y_{\varphi} = 0 \). This reveals their key weakness: such regularization imposes a strong inductive bias toward scenarios with small treatment effects. As a result, naively applying these approaches can degrade performance when this assumption is violated.
\section{Theoretical Analysis}
\label{sec:theory}
% \vspace{-.05in}
In this section, we analyze the statistical properties of the H-learner. We focus on the linear setup, where the H-learner admits a closed-form solution. We then characterize conditions under which H-learner achieves strictly lower prediction risk than either estimator alone. Extending to nonlinear function classes does not yield a tractable closed form, so we defer to experiments in Section~\ref{sec:experiments} for empirical validation. All proofs are deferred to Appendix~\ref{sec: add_theory}.

\subsection{Closed-Form Solution of the H-Learner in Linear Models}
\label{subsec:theory_linear}
% \vspace{-.05in}
Throughout this section, we use $X \in \mathbb{R}^{n \times d}$ to denote the covariate matrix and $X_1 \in \mathbb{R}^{n_1 \times d}$ and $X_0 \in \mathbb{R}^{n_0 \times d}$ denote the submatrices of $X$ corresponding to treated and control groups. Let $\tau(x)$ denote the true CATE, and define $\theta^\star = \arg\min_{\theta} \,\mathbb{E}\big[(x^\top \theta - \tau(x))^2\big]$ as the true treatment effect parameter in the linear $\tau(x)$ case, or the best linear projection of $\tau(x)$ onto the span of $x$ in the nonlinear case over the feature space. We denote by $\hat\theta_{\mathrm{ind}}$, $\hat\theta_{\mathrm{dir}}$, and $\hat\theta_{H}$ the OLS parameter estimates produced by the indirect learner, direct learner, and H-learner, respectively.
% , with estimation errors $e_{\mathrm{ind}} = \hat\theta_{\mathrm{ind}} - \theta^\star$ and $e_{\mathrm{dir}} = \hat\theta_{\mathrm{dir}} - \theta^\star$. We further define their biases $b_{\mathrm{ind}} = \mathbb{E}[e_{\mathrm{ind}}]$ and $b_{\mathrm{dir}} = \mathbb{E}[e_{\mathrm{dir}}]$, and their variances $\Sigma_{\mathrm{ind}} = \operatorname{Var}(e_{\mathrm{ind}})$, $\Sigma_{\mathrm{dir}} = \operatorname{Var}(e_{\mathrm{dir}})$ and $\Sigma_{\mathrm{ind,dir}} = \operatorname{Cov}(e_{\mathrm{ind}}, e_{\mathrm{dir}}).$ The mean squared error (MSE) can be expressed as $\mathrm{MSE}_{\mathrm{ind}} = \mathbb{E}[\lVert e_{\mathrm{ind}}\rVert_2^2]$, 
% $\mathrm{MSE}_{\mathrm{dir}} = \mathbb{E}[\lVert e_{\mathrm{dir}}\rVert_2^2]$.
For ease of notation, let $G_1 = X_1^\top X_1, \ G_0 = X_0^\top X_0, \ G = X^\top X$.

% \textbf{Assumption 5.1.} We assume covariates are isotropic, i.e., $X_1^\top X_1 = c_1 I_d, \ X_0^\top X_0 = c_0 I_d, \ X^\top X = c I_d$.

\textbf{Theorem 5.1} \textit{(Exact H-learner).}
Under the linear setup, the H-learner OLS estimator admits a closed-form solution given by a matrix-weighted combination of the indirect and direct OLS estimators.
\begin{align*}
\hat{\theta}_H = (I - W)\hat{\theta}_{\mathrm{ind}} + W\hat{\theta}_{\mathrm{dir}},
\end{align*}

\vspace{-0.1in}

where $W = \lambda A \big[(1 - \lambda) I + \lambda A\big]^{-1}, \ A := (G_1^{-1} + G_0^{-1})\, G.$

Theorem 5.1 shows that the H-learner forms a matrix-weighted linear combination with a weight that depends on the regularization parameter $\lambda$ and the geometry encoded in the matrix $A$. When $\lambda = 0$, the H-learner reduces to the indirect learner, and when $\lambda = 1$, it equals the direct learner. In general, this weighting operates in a direction-wise manner across the feature space. Let $\{\mu_i\}_{i=1}^d$ be the eigenvalues of $A$. Then the eigenvalues of $W$ are $\left\{\frac{\lambda \mu_i}{(1 - \lambda) + \lambda \mu_i}\right\}_{i=1}^d$, which are strictly increasing in both $\lambda$ and $\mu_i$. This highlights the adaptivity of the H-learner: Under poor overlap or in unbalanced settings with a small treatment arm, the eigenvalues of $A$ tend to increase, as the Gram matrix $G_1$ or $G_0$ corresponding to the smaller treatment group becomes ill-conditioned. Consequently, the eigenvalues of $W$ increase, shifting more weight toward the direct approach, where indirect learners are more susceptible to bias induced by
separate regularization.

We next analyze how this interpolation translates into improved prediction risk through a bias–variance decomposition.

\subsection{Bias–Variance Analysis of the H-Learner}
\label{subsec:theory_mse}
Let $e_{\mathrm{ind}} = \hat\theta_{\mathrm{ind}} - \theta^\star$ and
$e_{\mathrm{dir}} = \hat\theta_{\mathrm{dir}} - \theta^\star$
denote the estimation errors of the indirect and direct learners, respectively.
We then define their biases
$b_{\mathrm{ind}} = \mathbb{E}[e_{\mathrm{ind}}]$ and
$b_{\mathrm{dir}} = \mathbb{E}[e_{\mathrm{dir}}]$,
and their covariance matrices
$\Sigma_{\mathrm{ind}} = \operatorname{Var}(e_{\mathrm{ind}})$,
$\Sigma_{\mathrm{dir}} = \operatorname{Var}(e_{\mathrm{dir}})$,
$\Sigma_{\mathrm{ind,dir}} = \operatorname{Cov}(e_{\mathrm{ind}}, e_{\mathrm{dir}})$.
The mean squared error (MSE) can be expressed as
$\mathrm{MSE}_{\mathrm{ind}} = \mathbb{E}[\|e_{\mathrm{ind}}\|_2^2]$
and
$\mathrm{MSE}_{\mathrm{dir}} = \mathbb{E}[\|e_{\mathrm{dir}}\|_2^2]$.

To facilitate the analysis, we start by focusing on the scalar path $W=\omega I$, which leads to closed-form conditions under which the H-learner outperforms both endpoints. Since $\min_W \mathrm{MSE}_H(W)\le \min_{\omega\in[0,1]}\mathrm{MSE}_H(\omega I)$, any improvement achieved along this scalar path provides a sufficient condition, as the full matrix optimization yields no higher risk and thus will also improve upon both endpoints.

\textbf{Assumption 5.2.} We assume that the indirect and direct learners are trained via cross-fitting (e.g., on independent folds), so that $\Sigma_{\mathrm{ind,dir}} = 0.$

\textbf{Theorem 5.3} \textit{(MSE expansion and optimal weight).}
Under Assumptions~5.2, the MSE of the H-learner estimator is 
\begin{multline*}
\mathrm{MSE}_{\mathrm{H}} = \mathbb{E}\big[\lVert \hat\theta_H - \theta^\star \rVert_2^2\big]
= (1-\omega)^2\,\mathrm{MSE}_{\mathrm{ind}} \\
+\, \omega^2\,\mathrm{MSE}_{\mathrm{dir}}
+\, 2\omega(1-\omega)\,D,
\end{multline*}

\vspace{-0.15in}

where $D = b_{\mathrm{ind}}^\top b_{\mathrm{dir}}$.

The optimal convex weight that minimizes $\mathrm{MSE}_{\mathrm{H}}$ is
\begin{align*}
\omega^* = \min\!\left\{\,1,\; \max\!\left\{\,0,\; \frac{\mathrm{MSE}_{\mathrm{ind}} - D}{\mathrm{MSE}_{\mathrm{ind}} + \mathrm{MSE}_{\mathrm{dir}} - 2D}\,\right\}\right\}.
\end{align*}

Theorem 5.3 decomposes the H-learner’s MSE into a weighted sum of the direct and indirect errors together, and characterizes the optimal weight $\omega^*$. Consequently, the H-learner strictly outperforms both learners whenever $0 < \omega^* < 1$. We next present two corollaries that provide sufficient conditions for this to hold.

\vspace{0.05in}
\textbf{Corollary 5.4.}
If $b_{\mathrm{ind}}^\top b_{\mathrm{dir}} < 0$, then $\mathrm{MSE}_{\mathrm{H}} < \min\!\left\{ \mathrm{MSE}_{\mathrm{ind}},\; \mathrm{MSE}_{\mathrm{dir}} \right\}$.

\textbf{Corollary 5.5.}
If $\|b_{\mathrm{ind}}\|_2 \ge \|b_{\mathrm{dir}}\|_2$ and $\operatorname{tr}(\Sigma_{\mathrm{dir}}) > b_{\mathrm{dir}}^\top ( b_{\mathrm{ind}} - b_{\mathrm{dir}} )$, then $\mathrm{MSE}_{\mathrm{H}} < \min\!\left\{ \mathrm{MSE}_{\mathrm{ind}},\; \mathrm{MSE}_{\mathrm{dir}} \right\}$.

\vspace{0.1in}

Corollary 5.4 shows that when the biases point in opposite directions, the H-learner benefits from their cancellation. When they point in the same direction, Corollary~5.5 makes explicit the bias--variance trade-off between indirect and direct learners. It implies that whenever the indirect learner has a larger bias and the direct learner’s variance is sufficiently large relative to this bias gap, then the H-learner strictly outperforms both individual estimators.

\begin{figure}
\centering
\includegraphics[width=3.2in]{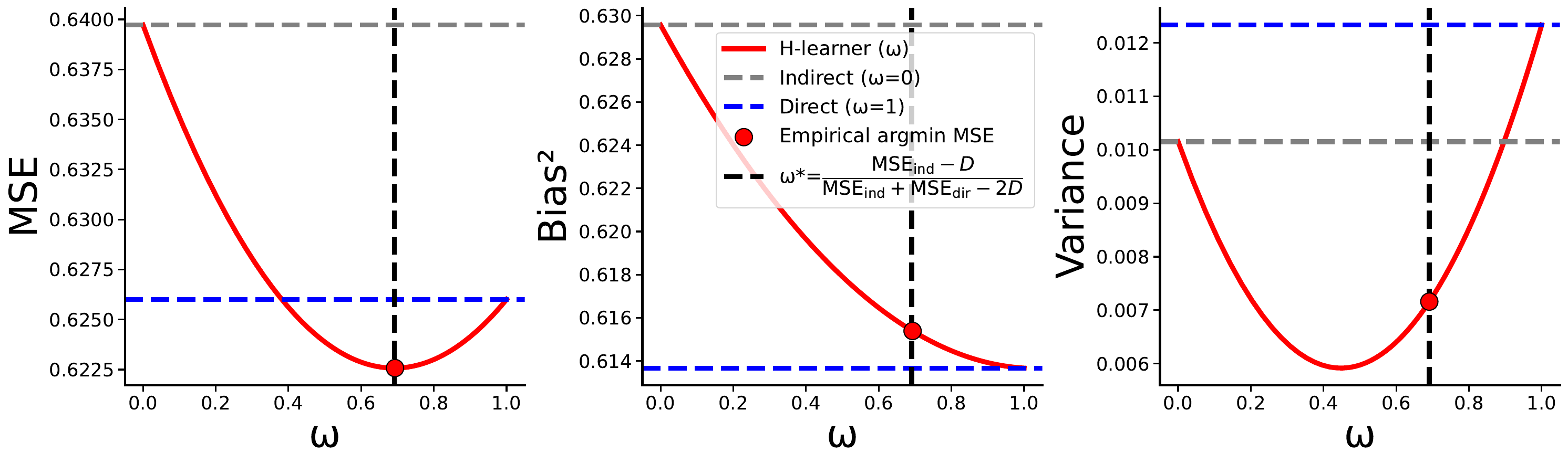} 
\caption{MSE and bias-variance decomposition of H-learner across convex weight $\omega$. Empirical optimum $\omega^\star$ aligns with the theoretical optimal point. H-learner with $\omega^\star$ achieves lower MSE than either baseline.
}
\label{fig:bias_variance}
\end{figure}

We further illustrate this with a 1D example in Figure~\ref{fig:bias_variance}. The plot shows the MSE of the H-learner as a function of the convex weight $\omega$, together with its decomposition into bias and variance. As expected, the indirect learner exhibits higher bias, while the direct learner suffers from higher variance. H-learner achieves strictly lower MSE than either baseline by reducing variance below both while maintaining a small level of bias.

\section{Experiments}
\label{sec:experiments}
% \vspace{-.05in}
In this section, we evaluate the performance of \method\ through semi-synthetic experiments and established benchmarks. Section~\ref{subsec:setup} outline the datasets and experimental setup. Section~\ref{subsec:synthetic} highlight the inconsistent performance between indirect and direct learners using custom-designed semi-synthetic setups and demonstrate the robustness of H-learner. In Section~\ref{subsec:benchmark}, we present results on widely used benchmark datasets, further highlighting the improved performance of \method\ over existing baselines.

\begin{figure*}[t]
  \centering
  \includegraphics[width=0.98\textwidth]{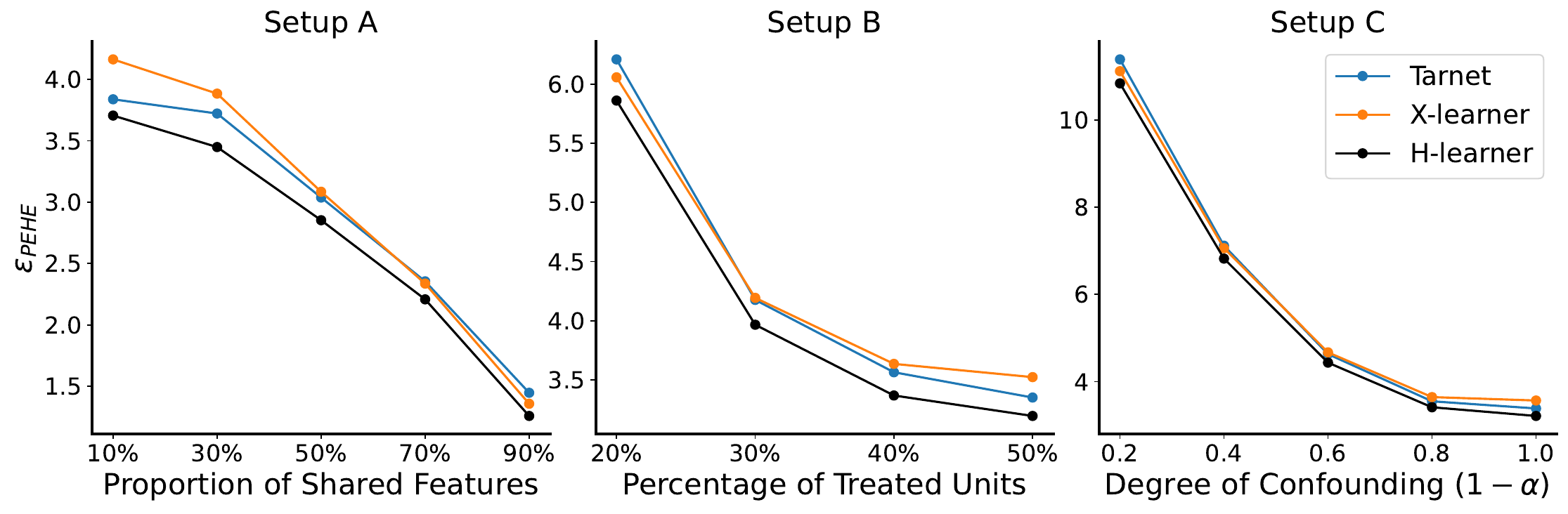}
    \caption{
    PEHE results for three semi-synthetic scenarios. In all setups, we observe performance trade-offs between TARNet (indirect learner) and X-learner (direct learner), while the proposed H-learner consistently outperforms both counterparts.}
  \label{fig:synthetic1}
\end{figure*}

\subsection{Setup}
\label{subsec:setup}
\textbf{Datasets.} We evaluate performance using two widely adopted benchmark datasets: IHDP and ACIC 2016, both commonly used in the literature on heterogeneous treatment effect estimation \citep{shalit2017estimating, alaa2017bayesian,
shi2019adaptingneuralnetworksestimation, 
lee2020robustrecursivepartitioningheterogeneous,
curth2021nonparametric}. Following prior work, we use a 63/27/10 train/validation/test split for both datasets.

\textbf{IHDP.} The IHDP dataset is based on covariates from the Infant Health and Development Program, a randomized study assessing the effect of specialist home visits on cognitive outcomes for premature infants \citep{hill2011bayesian}. Confounding and treatment imbalance are introduced by removing a subset of treated units. The resulting dataset includes 747 observations with 25 covariates. Following prior work, we evaluate performance by averaging over 1,000 realizations from setting “A” implemented in the NPCI package \citep{dorie2016npci}.

\textbf{ACIC 2016.} The ACIC 2016 datasets are derived from the Collaborative Perinatal Project and were introduced as part of the Atlantic Causal Inference Competition \citep{dorie2019automated}. This benchmark provides a comprehensive evaluation that varies in functional complexity, confounding, overlap, and treatment effect heterogeneity. It includes 77 distinct DGPs, with 4,802 observations and 79 covariates. Following \citep{lee2020robustrecursivepartitioningheterogeneous}, we remove rows where the POs below the 1st and above the 99th percentile to eliminate outliers and evaluate performance by averaging over 5 random realizations per setting.

% \textbf{Implementations.} To implement \method\, we choose TARNet as the backbone architecture among indirect learners due to its greater flexibility and sample efficiency compared to the T-learner and S-learner. For direct learners, we use the pseudo-outcomes from the X-learner and DR-learner as regularization targets. In all experiments, we train models using $\lambda \in \{0, 0.1, 0.2, \dots, 0.9, 1\}$, and select the best-performing model based on the validation criterion described in Section~\ref{subsec:lamda_selection}. Additional implementation details including hyperparameters are provided in the supplement.

\textbf{Evaluations.} We evaluate performance using \textit{Precision in Estimation of Heterogeneous Effect} (PEHE) \citep{hill2011bayesian}. The PEHE loss is defined as $\epsilon_{\text{PEHE}} = \frac{1}{n} \mbox{$\sum_{i=1}^n$} (\widehat{\tau}(x_i) - \tau(x_i))^2$, which corresponds to the mean squared error in estimating the CATE. We report performance on both in-sample (training and validation) and out-of-sample (test) data. Note that in-sample evaluation is non-trivial, as only the factual outcomes are observed during training, while the true CATE $\tau(x)$ is never observed.

\subsection{Semi-Synthetic Experiments}
\label{subsec:synthetic}
Real-world datasets often lead to inconsistent performance between direct and indirect learners due to a mixture of underlying factors. To disentangle these effects, we design semi-synthetic DGPs that isolate and examine three key sources of variation: (A) Relative complexity of POs and CATE, (B) Imbalance between treated and control units, and (C) Strength of confounding.

\textbf{Experiment setup.} In all setups, we use covariates from the IHDP dataset ($n=747, \ d=25$) but design custom response surfaces as follows. 

Let \( S_0 \) and \( S_1 \) denote sets of randomly selected features used to construct \( \mu_0(x) \) and \( \mu_1(x) \), respectively, and let \( S = S_0 \cup S_1 \). We then simulate the response surfaces and the treatment assignment as follows:

\vspace{-0.1in}

\begin{equation*}\label{eq:dgp}
\begin{aligned}
Y_i =
2 \ &\mbox{$\sum_{\substack{j \in S_{T_i}}}$} X_j
+ 2 \ \mbox{$\sum_{\substack{j \in S_{T_i}}}$} X_j^2
+ \mbox{$\sum_{\substack{j,k \in S_{T_i}}}$} X_j X_k
+ \varepsilon_i,\\[3pt]
&\mathbb{P}(T_i = 1 \mid X_i) =
\sigma\!\left( \alpha \cdot \mbox{$\sum_{\substack{j \in S}}$} \ \beta_j X_j \right),
\end{aligned}
\end{equation*}

% \vspace{-0.12in}

where \( \varepsilon_i, \beta_j \sim \mathcal{N}(0, 1) \), \( T_i \sim \text{Bernoulli}(\mathbb{P}(T_i = 1 \mid X_i)) \), and \( \sigma(\cdot) \) denotes the sigmoid function.

In Setup A, we vary the relative complexity of the CATE and PO functions by changing the proportion of shared features between \( S_0 \) and \( S_1 \) in \(\{10\%, 30\%, 50\%, 70\%, 90\%\}\), where a greater number of shared features implies a simpler CATE function. In Setup B, we vary group imbalance by adjusting the proportion of treated units in the dataset to \(\{20\%, 30\%, 40\%, 50\%\}\), where a higher percentage indicates more balanced treatment and control groups. In Setup C, we introduce varying levels of confounding in the treatment assignment function by adjusting \( \alpha \in \{0, 0.2, 0.4, 0.6, 0.8\} \), where larger values of \( \alpha \) correspond to stronger confounding. All simulations are averaged over 80 runs, with features in \( S_0 \) and \( S_1 \) randomly selected in each run.
% In Setup A, we vary the relative complexity of the CATE and PO functions by varying the proportion of shared features between $S_0$ and $S_1$
% in \{10\%, 30\%, 50\%, 70\%, 90\%\}, where increasing shared features implies a simpler CATE functions. In Setup B, we vary the group imbalance by changing the proportion of treated units in the dataset to \{20\%, 30\%, 40\%, 50\%\}, where a higher percentage indicates a more balanced treatment-control groups. In Setup C, we introduce varying levels of confounding in the treatment assignment function by adjusting \( \alpha \in \{0, 0.2, 0.4, 0.6, 0.8\} \), where larger values of \( \alpha \) correspond to stronger confounding.

% In all setups, we fix the sizes of \( S_0 \) and \( S_1 \) to 10. To ensure that only one factor varies at a time, we set \( \alpha = 0 \) in Setups A and B, and fix the proportion of shared features at 40\% in Setups B and C. All simulations are averaged over 80 runs, with features in \( S_0 \) and \( S_1 \) randomly selected in each run.

% \begin{figure}
% \centering
% \includegraphics[width=0.8\columnwidth]{Figures/synthetic_fig2_update.pdf} 
% % \vspace{-.1in}
% \caption{Performance of H-learner across different $\lambda$ values for Setup A.}
% \label{fig:synthetic2}
% \end{figure}

\textbf{Results.} Figure~\ref{fig:synthetic1} presents results from the three simulated scenarios. We show the performance of the H-learner using X-learner pseudo-outcomes in its first stage and compare it against its constituent components: the X-learner (direct learner) and TARNet (indirect learner). Results using alternative learners are provided in Appendix \ref{sec: add_syn}.

\textbf{Setup A} shows the performance trade-off through the \textit{relative complexity of the POs and the CATE}. As the proportion of shared features increases, the PO functions become more similar, which leads to a simpler CATE surface. Correspondingly, we observe that the X-learner only outperforms TARNet when the proportion of shared features is high, as its constructed pseudo-outcomes target a simpler function, but underperforms when the shared feature proportion is low.

\begin{figure}
\centering
\includegraphics[width=0.85\columnwidth]{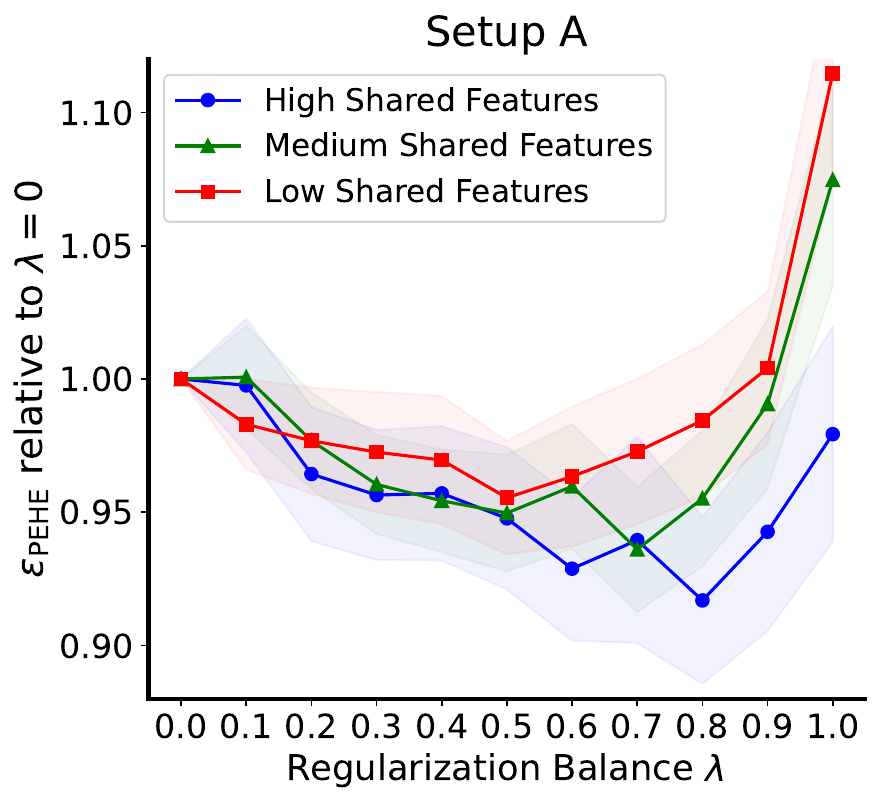} 
\caption{Performance of H-learner across different $\lambda$ values for Setup A. Optimal performance is consistently achieved at intermediate $\lambda$ values.}
% \vspace{-.14in}
\label{fig:synthetic2}
\end{figure}

\textbf{Setup B} demonstrates the performance trade-off 
driven by the \textit{treatment–control imbalance}. As the proportion of treated units decreases, the imbalance in sample sizes between the treatment and control groups becomes more severe. This imbalance disproportionately affects indirect learners, as their separate outcome models are trained on increasingly unbalanced samples and are more susceptible to ``regularization-induced confounding''. Empirically, TARNet performs better under balanced treatment but underperforms when only 20\% of units receive treatment.

\textbf{Setup C} demonstrates the performance trade-off driven by the \textit{strength of confounding}. As the degree of confounding increases, the feature distributions differ more significantly across treatment groups. This poses greater challenges for indirect learners, as their separate outcome models are trained on increasingly divergent data distributions and suffer from reduced sample efficiency. As expected, TARNet performs well only under weaker confounding, but fails to match the X-learner when confounding is stronger.

In contrast, across all varying setups, the H-learner consistently outperforms both its direct and indirect counterparts by adjusting its inductive bias to better align with the underlying data, demonstrating robustness by unifying the strengths of both learners.

\textbf{Effect of regularization parameter $\lambda$}. We further investigate the performance of H-learner across different $\lambda$ values. Figure~\ref{fig:synthetic2} shows how the test PEHE varies with $\lambda$ in Setup A, across different levels of feature sharing: low (10\%), medium (50\%), and high (90\%).

\textbf{Results.} Figure~\ref{fig:synthetic2} shows that the optimal value of $\lambda$ consistently lies between the two extremes—$\lambda = 0$ (indirect learner) and $\lambda = 1$ (direct learner)—further highlighting that combining both regularizations yields better performance than relying on either alone. Across different levels of feature sharing, the H-learner with the best-performing $\lambda$ consistently improves upon the weaker learner by at least 10\% and the stronger learner by at least 5\%. Moreover, the trend in the optimal $\lambda$ reflects the inductive bias inherent in the data: as more features are shared, the underlying CATE function becomes simpler and tends to favor direct regularization. We observe that the H-learner adapts to this inductive bias by yielding a larger optimal $\lambda$ as the number of shared features increases. Similar performance results in Setups B and C are provided in Appendix \ref{sec: add_syn}.

\subsection{Benchmark Evaluation}
\label{subsec:benchmark}
Here we present results on the IHDP and ACIC 2016 benchmark datasets, comparing the H-learner with baseline meta-learners and regularization strategies from the literature.

\begin{table}[t]
\centering
\caption{Within-sample and out-of-sample PEHE results on benchmark datasets IHDP and ACIC 2016. Entries marked with $^{\star}$ indicate statistically significant improvements over all baselines
based on one-sided paired t-tests.}
% \vspace{-0.05in}
\resizebox{\columnwidth}{!}{%
\begin{tabular}{llcccc}
\toprule
\multicolumn{2}{c}{} & \multicolumn{2}{c}{\textbf{IHDP}} & \multicolumn{2}{c}{\textbf{ACIC 2016}} \\
\cmidrule(lr){3-4} \cmidrule(lr){5-6}
\multicolumn{2}{c}{} 
& \makecell{In-sample\\$\sqrt{\epsilon_{\text{PEHE}}}$} 
& \makecell{Out-of-sample\\$\sqrt{\epsilon_{\text{PEHE}}}$} 
& \makecell{In-sample\\$\sqrt{\epsilon_{\text{PEHE}}}$} 
& \makecell{Out-of-sample\\$\sqrt{\epsilon_{\text{PEHE}}}$}  \\
\midrule
& T-learner & $0.97 \pm .02$ & $1.07 \pm .03$ & $2.08\pm .04$ &  $2.19 \pm .04$\\
& TARNet & $0.81 \pm .02$ & $0.93 \pm .03$ & $1.66 \pm .03$ & $1.77 \pm .04$\\
\midrule
& IPW-learner & $4.93 \pm .19$ & $4.95 \pm .20$ & $2.69\pm .07$ &  $2.68 \pm .07$\\
& DR-learner & $0.98 \pm .02$ & $1.08 \pm .02$ & $1.63 \pm .04$ & $1.75 \pm .04$\\
& X-learner & $0.79 \pm .01$ & $0.91 \pm .02$ & $1.58\pm .03$ &  $1.70 \pm .04$\\
% & R-learner & $1.34 \pm .03$ & $1.41 \pm .04$ & $1.64 \pm .04$ &  $1.75 \pm .05$\\
\midrule
& TARNet-WR & $1.19 \pm .03$ & $1.33 \pm .04$ & $1.89\pm .04$ &  $1.97 \pm .04$\\
& OffsetNet & $2.11 \pm .08$ & $2.18 \pm .09$ & $2.10 \pm .05$ & $2.15 \pm .05$\\
& FlexTENet & $1.18 \pm .03$ & $1.30 \pm .04$ & $1.81 \pm .04$ & $1.89 \pm .04$\\
% \midrule
% & Super-learner (DR) & $0.96 \pm .03$ & $1.06 \pm .03$ & $1.60 \pm .04$ &  $1.69 \pm .04$\\
% & Super-learner (X) & $0.94 \pm .03$ & $1.04 \pm .03$ & $1.60 \pm .04$ &  $1.69 \pm .04$\\
\midrule
& \method\ (DR) & $0.81 \pm .01$ & $0.91 \pm .02$ & $1.57 \pm .03$& $1.68 \pm .04$\\
& \method\ (X) 
& $\bm{0.78 \pm .01}^{\star}$ 
& $\bm{0.88 \pm .02}^{\star}$ 
& $\bm{1.56 \pm .03}^{\star}$ 
& $\bm{1.67 \pm .04}^{\star}$ \\
\bottomrule
\end{tabular}
}
\label{tab:real_performance}
\end{table}

\textbf{Baselines.} We consider T-learner and TARNet as baselines for indirect learners, and X-learner, DR-learner, and IPW-learner as baselines for direct learners. Additionally, we include three state-of-the-art regularization strategies: (1) \textit{Weight Regularization}: TARNet-WR, a variant of TARNet with an explicit regularization that penalizes differences between the weights of the two output heads \citep{hahn2018regularization, curth2021inductive}. (2) \textit{Reparametrization:} OffsetNet, a neural network--based indirect learner that achieves implicit regularization through the reparameterization \( \mu_1(x) = \mu_0(x) + \tau(x) \), enabling direct estimation of \( \tau(x) \) as an offset from \( \mu_0(x) \) \citep{imai2013estimating, curth2021inductive}. (3) \textit{Structural Regularization:} FlexTENet, a neural network architecture that imposes regularization through flexible feature sharing across layers \citep{curth2021inductive}.

\textbf{Results.} As shown in Table~\ref{tab:real_performance}, \method\ outperforms all other baselines in terms of PEHE on both datasets. The best results are achieved when \method\ constructs X-learner pseudo-outcomes in its first stage.

\begin{figure}
\centering
\includegraphics[width=3.3in]{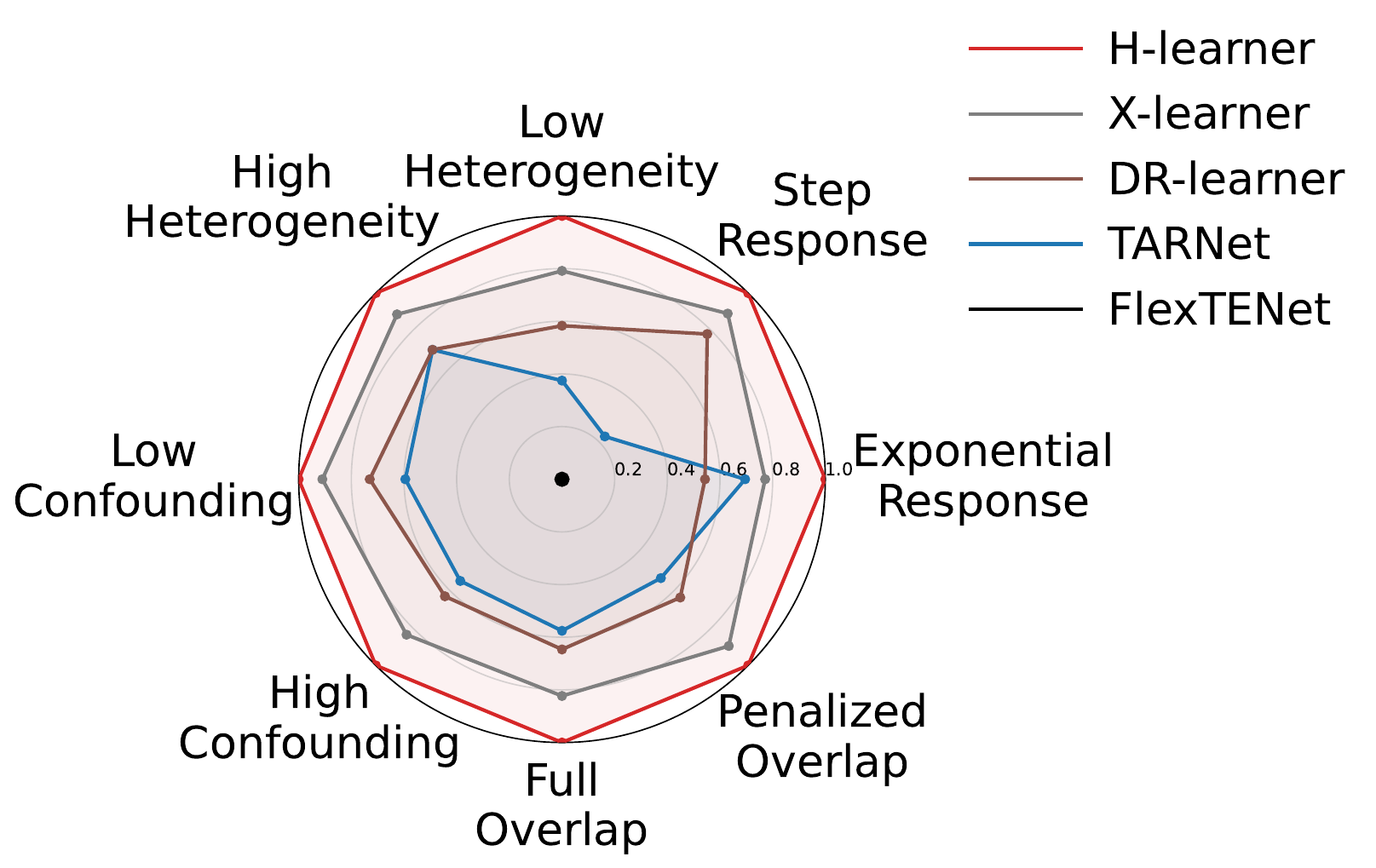} 
%\vspace{-.1in}
\caption{Radar chart of ACIC performance by DGP characteristics. PEHE is min–max normalized so that higher values indicate better performance, with 1 denoting the best. Only the top method per class is shown. H-learner performs best across all settings.}
\vspace{-0.05in}
\label{fig:acic_radar}
\end{figure}

\textbf{Comparison to direct and indirect learners:}
Among indirect learners, TARNet outperforms the T-learner due to its shared representation and improved sample efficiency. Among direct learners, the X-learner outperforms both the IPW-learner and DR-learner, which suffer from higher variance. Notably, \method\ consistently outperforms both its indirect and direct counterparts on both datasets when constructing pseudo-outcomes from either the DR-learner or X-learner in the first stage.

\textbf{Comparison to other regularization strategies:} Note that all other existing regularization strategies fail to perform well on these benchmarks. As discussed in Section~\ref{subsec:related2}, these strategies rely on the assumption that reparameterization or sharing of weights simplifies the learning problem—an assumption that only holds when the treatment effect is small. Since the assumption fails to hold in practical scenarios, using these strategies can actually harm performance. For example, TARNet with weight regularization performs worse than the standard TARNet model. This further highlights the robustness of \method\ in handling diverse data conditions without relying on any structural assumptions.

\textbf{Performance breakdown by DGP characteristics:} We further analyze performance on ACIC 2016 by decomposing results along specific DGP characteristics, as summarized in Figure \ref{fig:acic_radar}. The H-learner consistently achieves the best performance across a wide range of settings, including variations in response models, treatment effect heterogeneity, confounding strength, and covariate overlap. Complete results are provided in Appendix~\ref{sec:add_acic}.
\section{Discussion}
\label{sec:discussion}
% \vspace{-.05in}
Meta-learning approaches remain a prominent strategy for estimating CATE due to their intuitive design and strong empirical performance. However, achieving robustness across diverse DGPs requires a deeper understanding of how different meta-learners encode inductive biases through their regularization strategies. Indirect learners regularize models for POs, while direct learners directly regularize the CATE function by constructing pseudo-outcome estimators. These differing inductive biases make certain strategies more effective under specific DGPs while failing on others.

% Indirect learners regularize models for POs, while direct learners directly regularize the CATE function by constructing pseudo-outcome estimators. 

\method\ introduces a generalized regularization strategy that interpolates between indirect and direct regularizations while adaptively tuning the appropriate strength of each. This adaptivity enables \method\ to adjust its inductive bias to better match the underlying structure of the data without requiring prior knowledge of the DGPs. Experiments show that \method\ effectively overcomes the trade-off between indirect and direct learners, and achieving robust performance.

H-learner also offers several promising directions for future extensions. Potential directions include integrating H-learner with more complex causal models, such as those involving hidden confounding or multi-level treatments. Further, our experiments are currently limited to neural network implementations, extending \method\ to other ML models is left for future work.

% Despite improved performance, our study has limitations. One limitation of \method\ is its increased computational cost, as selecting the optimal regularization level requires training multiple models across different $\lambda$ values and selecting the best based on validation performance. Additionally, our experiments are currently limited to neural network implementations, extending \method\ to other machine learning models is left as future work.

\section*{Acknowledgements}

This work was supported by a Hellman Fellowship award from the Society of Hellman Fellows and the Evergreen Foundation.

\clearpage
\bibliography{references}
\bibliographystyle{plainnat}

\onecolumn
\clearpage
\appendix
\section{Implementations Details}
\label{sec:implementations}
\setcounter{equation}{0}
\renewcommand{\theequation}{\arabic{equation}}

\subsection{Pseudo-algorithm of H-learner}
Below, we present the pseudo-code for the H-learner.

\begin{minipage}{0.55\linewidth}
\begin{algorithm}[H]
{
    \SetKwInOut{Input}{{\footnotesize Input}}
    \SetKwInOut{Output}{{\footnotesize Output}}
    
    \Input{\, \mbox{\scriptsize $\mathcal{D} = \{(X_i, T_i, Y_i)\}_{i=1}^n$}}
    \Output{\, \mbox{\scriptsize $\widehat{\theta}(x) = \widehat{f}_1(x) - \widehat{f}_0(x)$}}
    
    {\it Split \mbox{\scriptsize $\mathcal{D}$} into \mbox{\scriptsize $\mathcal{D}_{1}$} and \mbox{\scriptsize $\mathcal{D}_{2}$}} \;
    
    \textbf{Stage 1: Pseudo-outcome construction} \;
    Estimate \mbox{\scriptsize $\widehat{\varphi}=(\widehat{\pi}, \widehat{\mu}_0, \widehat{\mu}_1)$} using \mbox{\scriptsize $\mathcal{D}_{1}$} \;
    Construct \mbox{\scriptsize ${Y}_{\varphi}$} via the options in Table~\ref{table:1} \;
    
    \textbf{Stage 2: Training the H-learner} \;
    Train \mbox{\scriptsize $\widehat{f}_0, \widehat{f}_1$} on \mbox{\scriptsize $\mathcal{D}_{2}$} using the loss in (\ref{hlearner:loss}) \;
    
    \Return \mbox{\scriptsize $\widehat{\theta}(x) = \widehat{f}_1(x) - \widehat{f}_0(x)$}
}
\caption{{\footnotesize {\bf H-learner}}}
\end{algorithm}
\end{minipage}

\subsection{Model Architectures and Training Details}
We implement all models following the architecture used in \citep{shalit2017estimating}. Specifically, the representation network of H-learner consists of three fully connected layers, each with 200 units and exponential linear unit (ELU) activations. Each output head comprises two additional dense layers with 100 units, followed by a final prediction layer. We implement similar architectures for the other meta-learners to ensure a fair comparison. All models are trained for 1000 epochs using the AdamW optimizer with mini-batches of size 100 and a cosine annealing learning rate schedule. We perform a grid search over learning rates \{0.0001, 0.0005, 0.001\}, and checkpoint achieving the lowest validation loss is used for final evaluation. We detail the validation criteria in the following section.

While we propose the H-learner with sample splitting using $\mathcal{D}_{1}$ and $\mathcal{D}_{2}$ for first and second stages as is commonly done in the literature, we found that using the full dataset for both stages often yields better empirical performance, particularly in small-sample settings. Similar observations have been made in \citet{curth2021nonparametric}, where sample splitting is then omitted. We therefore do not apply sample splitting and instead train both stages using the full dataset $\mathcal{D}$. Experiments were conducted on machine with an AMD EPYC 7543 32-Core Processor.

\subsection{Hyperparameter Tuning}
Hyperparameter tuning in CATE estimation is inherently challenging, as the true CATE is never observed, making standard supervised validation infeasible. To address this, we impute the unobserved counterfactual outcome using first-stage estimates \(\hat{\mu}_0(x)\) and \(\hat{\mu}_1(x)\), and use the following proxy loss on the validation set to select the best model checkpoint and hyperparameter:
\[
\widehat{\text{PEHE}}(f_0, f_1) = \frac{1}{n} \sum_{i=1}^n \left[ \left(f_1(x_i) - f_0(x_i)\right) - \left( t_i (y_i - \hat{\mu}_0(x_i)) + (1 - t_i)(\hat{\mu}_1(x_i) - y_i) \right) \right]^2.
\]
However, the above mechanism cannot be used to select the regularization strength \(\lambda\), since imputing the unobserved counterfactual outcomes using first-stage estimates produces X-learner pseudo-outcomes $Y_{\varphi,i} = T_i (Y_i - \hat{\mu}_0(X_i)) + (1 - T_i)(\hat{\mu}_1(X_i) - Y_i)$. As a result, when H-learner constructs X-learner pseudo-outcomes in its first stage, this validation loss will be biased toward favoring no indirect regularization (i.e., \(\lambda = 1\)).

To address this issue, we instead estimate the POs by training outcome models on the \textbf{validation set}, yielding \(\check{\mu}_0(X_i)\) and \(\check{\mu}_1(X_i)\). We then select the value of \(\lambda\) that minimizes the following validation loss: $\frac{1}{n} \sum_{i=1}^n \left[ \left(f_1(x_i) - f_0(x_i)\right) - \left( t_i (y_i - \check{\mu}_0(x_i)) + (1 - t_i)(\check{\mu}_1(x_i) - y_i) \right) \right]^2.$ Similar strategies have also been employed in prior works \citep{alaa2019validating, mahajan2022empirical}. We compare the $\lambda$ chosen by our validation-based strategy with the ground-truth optimal $\lambda^\ast$ (minimizing test PEHE) in Table~\ref{tab:lambda_selection}, showing that the validation-selected $\lambda$ closely tracks the optimal value. Note that all validations are performed using the same validation set as other meta-learners, ensuring \method\ uses the same amount of information for fair comparisons.

\begin{table}[h]
\centering
\begin{tabular}{c|c|c}
\hline
\textbf{Feature Overlap Ratio} & \textbf{$\lambda^\ast$ (Test PEHE)} & \textbf{$\lambda$ (Validation X-score)} \\
\hline
0.1 & 0.49 & 0.52 \\
0.3 & 0.54 & 0.57 \\
0.5 & 0.57 & 0.58 \\
0.7 & 0.57 & 0.58 \\
0.9 & 0.67 & 0.62 \\
\hline
\end{tabular}
\caption{Comparison between the optimal $\lambda^\ast$ minimizing test PEHE and the $\lambda$ selected using the X-score on the validation set in the first synthetic setup of Section \ref{subsec:synthetic}.}
\label{tab:lambda_selection}
\end{table}

\section{Proof and Additional Theoretical Analysis}
\label{sec: add_theory}
\subsection{Complete Proof for the H-Learner Solution in Linear Models}
\textbf{Setup.}
Let $X \in \mathbb{R}^{n \times d}$ denote the covariate matrix, $T \in \{0,1\}^n$ denote the treatment vector, and $Y \in \mathbb{R}^n$ denote the observed outcomes. Let $X_1 \in \mathbb{R}^{n_1 \times d}$ and $X_0 \in \mathbb{R}^{n_0 \times d}$ denote the submatrices of $X$ corresponding to treated ($T=1$) and control ($T=0$) units, and similarly $Y_1 \in \mathbb{R}^{n_1}$ and $Y_0 \in \mathbb{R}^{n_0}$. Let $Y_\phi \in \mathbb{R}^n$ be the pseudo-outcomes. For ease of notation, let $G_1 = X_1^\top X_1, \ G_0 = X_0^\top X_0, \ G = X^\top X, \ S_1 = X_1^\top Y_1, \ S_0 = X_0^\top Y_0, \ S_\phi = X^\top Y_\phi$.

% Under the assumption that the covariates are isotropic:
% \[
% X_1^\top X_1 = c_1 I_d,\qquad
% X_0^\top X_0 = c_0 I_d,\qquad
% X^\top X = c I_d, \quad \textit{where} \ \ c = c_0 + c_1.
% \]

\textbf{Indirect and direct estimators.} The indirect learner fits separate linear models to both groups:
\[
\hat{\theta}^{(1)}_{\mathrm{ind}} = (X_1^\top X_1)^{-1} X_1^\top Y_1 = G_1^{-1} S_1,\qquad
\hat{\theta}^{(0)}_{\mathrm{ind}} = (X_0^\top X_0)^{-1} X_0^\top Y_0= G_0^{-1} S_0,\qquad
\hat{\theta}_{\mathrm{ind}} = \hat{\theta}^{(1)}_{\mathrm{ind}} - \hat{\theta}^{(0)}_{\mathrm{ind}}
\]

The direct estimator from the pseudo-outcome regression is
\[
\hat{\theta}_{\mathrm{dir}} = (X^\top X)^{-1} X^\top Y_\phi = G^{-1} S_\phi
\]

\textbf{H-learner.} We study the H-learner setup by modeling $f_1$ and $f_0$ as linear models with parameters $\theta_1, \theta_0 \in \mathbb{R}^d$:
\[
f_1(x) = \theta_1^\top x, \qquad
f_0(x) = \theta_0^\top x, \qquad
f_1(x) - f_0(x) = (\theta_1 - \theta_0)^\top x = \hat{\theta}_H^\top x
\]

We minimize the H-learner objective
\[
L(\theta_1,\theta_0)
= (1-\lambda)\big(\|Y_1 - X_1\theta_1\|_2^2 + \|Y_0 - X_0\theta_0\|_2^2\big)
+ \lambda\,\|X(\theta_1-\theta_0) - Y_\phi\|_2^2
\]

Taking gradients and setting to zero yields
\begin{align*}
\frac{\partial L}{\partial \theta_1}
= -2(1-\lambda) X_1^\top (Y_1 - X_1 \theta_1)
   + 2\lambda X^\top (X (\theta_1 - \theta_0) - Y_\phi) = 0\\
\frac{\partial L}{\partial \theta_0}
= -2(1-\lambda) X_0^\top (Y_0 - X_0 \theta_0)
   - 2\lambda X^\top (X (\theta_1 - \theta_0) - Y_\phi) = 0
\end{align*}

Rearrange:
\begin{align}
(1-\lambda) G_1 \theta_1 + \lambda G (\theta_1 - \theta_0)
&= (1-\lambda) S_1 + \lambda S_\phi
\label{eq:b11}\\
(1-\lambda) G_0 \theta_0 - \lambda G (\theta_1 - \theta_0)
&= (1-\lambda) S_0 - \lambda S_\phi
\label{eq:b12}
\end{align}

Multiply \eqref{eq:b11} by $G_1^{-1}$ and \eqref{eq:b12} by $G_0^{-1}$:
\begin{align*}
(1-\lambda)\theta_1 + \lambda G_1^{-1} G (\theta_1 - \theta_0)
&= (1-\lambda) G_1^{-1} S_1 + \lambda G_1^{-1} S_\phi, \\
(1-\lambda)\theta_0 - \lambda G_0^{-1} G (\theta_1 - \theta_0)
&= (1-\lambda) G_0^{-1} S_0 - \lambda G_0^{-1} S_\phi
\end{align*}

Subtract the second from the first:
\[
\big[(1-\lambda) I + \lambda (G_1^{-1} + G_0^{-1}) G \big] (\theta_1 - \theta_0)
= (1-\lambda)(G_1^{-1} S_1 - G_0^{-1} S_0) + \lambda (G_1^{-1} + G_0^{-1}) S_\phi
\]

Then
\[
\big[(1-\lambda) I + \lambda A \big] \hat{\theta}_H
= (1-\lambda)\hat{\theta}_{\text{ind}} + \lambda A \hat{\theta}_{\text{dir}},
\quad \text{where } A = (G_1^{-1} + G_0^{-1}) G
\]

Hence,
\begin{align*}
\hat{\theta}_H
&= \big[(1-\lambda) I + \lambda A \big]^{-1}
   \big[(1-\lambda)\hat{\theta}_{\text{ind}} + \lambda A \hat{\theta}_{\text{dir}}\big] \\
&= (1-\lambda)\big[(1-\lambda) I + \lambda A \big]^{-1}\hat{\theta}_{\text{ind}}
   + \lambda A\big[(1-\lambda) I + \lambda A \big]^{-1} \hat{\theta}_{\text{dir}}
\end{align*}

Therefore,
\[
\hat{\theta}_H
= (I - W)\hat{\theta}_{\text{ind}} + W\hat{\theta}_{\text{dir}},
\quad
W = \lambda A \big[(1-\lambda) I + \lambda A \big]^{-1}
\]

\subsection{Bias–Variance Analysis of the H-Learner Estimator}
\subsubsection{Equivalence Between Parameter-Space MSE and CATE Prediction Error}

\paragraph{Assumption (Isotropy).} 
Throughout this analysis, we assume the covariates are isotropic, i.e.,
\[
\Sigma_X := \mathbb{E}[X X^\top] = c I_d, \text{ for some constant } c > 0.
\]

\textbf{Case 1:} When the true CATE function is linear $\tau(x) = x^\top \theta^\star$, then for any estimator $\hat{\theta}$, the CATE prediction error is
\begin{align}
\mathbb{E}\!\left[(\hat{\tau}(X) - \tau(X))^2\right] 
&= \mathbb{E}\!\left[(X^\top\hat{\theta} - X^\top\theta^\star)^2\right] 
= \mathbb{E}\!\left[(X^\top(\hat{\theta}-\theta^\star))^2\right] \nonumber \\[4pt]
&= (\hat{\theta}-\theta^\star)^\top\,\mathbb{E}[XX^\top]\,(\hat{\theta}-\theta^\star) 
= (\hat{\theta}-\theta^\star)^\top\,\Sigma_X\,(\hat{\theta}-\theta^\star)
\label{eq:cate-mse}
\end{align}
Under isotropy, $\Sigma_X =c I_d$, so \eqref{eq:cate-mse} becomes
\[
\mathbb{E}\!\left[(\hat{\tau}(X) - \tau(X))^2\right] 
= c\,\|\hat{\theta}-\theta^\star\|_2^2.
\]
Therefore, comparing CATE error is equivalent (up to the constant) to comparing their parameter-space errors:
\[
\mathbb{E}\!\left[(\hat{\tau}(X) - \tau(X))^2\right] 
\;\propto\; \mathbb{E}\!\left[\|\hat{\theta}-\theta^\star\|_2^2\right]
\]

\textbf{Case 2:} When the true CATE function $\tau(x)$ is not linear, let
\[
\theta^\star := \arg\min_{\theta} \ \mathbb{E}\big[(x^\top \theta - \tau(x))^2\big]
\]
be its best linear projection. Then for any estimator $\hat{\theta}$,
\begin{align*}
\mathbb{E}\big[(x^\top \hat{\theta} - \tau(x))^2\big]
&=
\mathbb{E}\big[(x^\top \theta^\star - \tau(x))^2\big]
+
\mathbb{E}\big[(x^\top \hat{\theta} - x^\top \theta^\star)^2\big] \\[4pt]
&=
\underbrace{\mathbb{E}\big[(x^\top \theta^\star - \tau(x))^2\big]}_{\text{approximation error (independent of } \hat{\theta}\text{)}} 
\;+\;
c \, \mathbb{E}\big[\|\hat{\theta} - \theta^\star\|_2^2\big].
\end{align*}

\vspace{-.1in}

Therefore, even in the nonlinear case, all estimator comparisons can be carried out entirely in parameter space via $\mathbb{E}\|\hat{\theta} - \theta^\star\|_2^2$. Hence, in the following sections, we compare estimators through their parameter-space MSE.

\subsubsection{Mean Squared Error Decomposition for the H-Learner}
Let $e_{\mathrm{ind}} = \hat{\theta}_{\mathrm{ind}} - \theta^\star, e_{\mathrm{dir}} = \hat{\theta}_{\mathrm{dir}} - \theta^\star$ denote the estimation errors of the indirect and direct estimators, denote the estimation errors of the indirect and direct estimators, where $\theta^\star \in \mathbb{R}^d$ is the true treatment-effect parameter in the linear CATE case, or the best linear projection as defined above in the nonlinear case.

Define their biases and covariances as
\[
b_{\mathrm{ind}} = \mathbb{E}[e_{\mathrm{ind}}], 
\qquad
b_{\mathrm{dir}} = \mathbb{E}[e_{\mathrm{dir}}],
\qquad
\Sigma_{\mathrm{ind}} = \operatorname{Var}(e_{\mathrm{ind}}),
\qquad
\Sigma_{\mathrm{dir}} = \operatorname{Var}(e_{\mathrm{dir}})
\]

Then their mean squared errors can be written as
\[
\mathrm{MSE}_{\mathrm{ind}} = \mathbb{E}\big[\|\hat{\theta}_{\mathrm{ind}} - \theta^\star\|_2^2\big] = \|b_{\mathrm{ind}}\|^2 + \operatorname{tr}(\Sigma_{\mathrm{ind}}),
\qquad
\mathrm{MSE}_{\mathrm{dir}} = \mathbb{E}\big[\|\hat{\theta}_{\mathrm{dir}} - \theta^\star\|_2^2\big] =  \|b_{\mathrm{dir}}\|^2 + \operatorname{tr}(\Sigma_{\mathrm{dir}})
\]

\textbf{H-learner.}
We begin by analyzing the MSE along the scalar path \( W = \omega I \), under which $\hat{\theta}_H$ forms a convex combination of the indirect and direct OLS estimators: $\hat{\theta}_H = (1 - \omega)\,\hat{\theta}_{\mathrm{ind}} + \omega\,\hat{\theta}_{\mathrm{dir}}.$ 

We then compute the bias of the H-learner.
\begin{align*}
b_H 
&= \mathbb{E}[\hat{\theta}_H - \theta^\star] \\
&= \mathbb{E}\left[(1-\omega)\,\hat{\theta}_{\mathrm{ind}} + \omega\,\hat{\theta}_{\mathrm{dir}} - \theta^\star\right] \\
&= (1-\omega)\,b_{\mathrm{ind}} + \omega\,b_{\mathrm{dir}}.
\end{align*}

Variance of the H-learner:
\begin{align*}
\operatorname{Var}(\hat{\theta}_H)
&= \operatorname{Var}\big((1-\omega)\,\hat{\theta}_{\mathrm{ind}} + \omega\,\hat{\theta}_{\mathrm{dir}}\big) \\
&= (1-\omega)^2 \operatorname{Var}(\hat{\theta}_{\mathrm{ind}}) + \omega^2 \operatorname{Var}(\hat{\theta}_{\mathrm{dir}}) 
  + 2\omega(1-\omega)\,\operatorname{Cov}(\hat{\theta}_{\mathrm{ind}}, \hat{\theta}_{\mathrm{dir}}).
\end{align*}

Therefore, the MSE of the H-learner is:
\begin{align*}
\mathrm{MSE}_H 
&= \mathbb{E}\!\left[\big\|\hat{\theta}_H-\theta^\star\big\|_2^2\right] \\
&= \|b_H\|_2^2 + \operatorname{tr}\!\big(\operatorname{Var}(\hat{\theta}_H)\big) \nonumber\\
&= \big\|(1-\omega)\,b_{\mathrm{ind}}+\omega\,b_{\mathrm{dir}}\big\|_2^2
   + \operatorname{tr}\!\Big((1-\omega)^2\Sigma_{\mathrm{ind}}
   + \omega^2\Sigma_{\mathrm{dir}}
   + 2\omega(1-\omega)\Sigma_{\mathrm{ind,dir}}\Big)\\
&= (1-\omega)^2\!\left(\|b_{\mathrm{ind}}\|_2^2+\operatorname{tr}(\Sigma_{\mathrm{ind}})\right)
   + \omega^2\!\left(\|b_{\mathrm{dir}}\|_2^2+\operatorname{tr}(\Sigma_{\mathrm{dir}})\right)
   + 2\omega(1-\omega)\!\left(b_{\mathrm{ind}}^\top b_{\mathrm{dir}}
   + \operatorname{tr}(\Sigma_{\mathrm{ind,dir}})\right)\\
&= (1-\omega)^2\,\mathrm{MSE}_{\mathrm{ind}}
   + \omega^2\,\mathrm{MSE}_{\mathrm{dir}}
   + 2\omega(1-\omega)\,D, \nonumber
\end{align*}
where $D = b_{\mathrm{ind}}^\top b_{\mathrm{dir}} + \operatorname{tr}\big(\Sigma_{\mathrm{ind,dir}}\big), \ \ \Sigma_{\mathrm{ind,dir}} = \operatorname{Cov}(\hat{\theta}_{\mathrm{ind}}, \hat{\theta}_{\mathrm{dir}})$

\subsubsection{When Does the H-Learner Strictly Improve MSE?}
We now analyze conditions that the optimal convex combination \(\hat{\theta}_H = (1-\omega)\,\hat{\theta}_{\mathrm{ind}} + \omega\,\hat{\theta}_{\mathrm{dir}}\)
achieve \emph{strictly smaller} MSE than either endpoint ($\omega = 0$ or $\omega = 1$).

From the quadratic form of the H-learner MSE, the optimal weight is
\[
    \omega^\star = \frac{\mathrm {MSE}_{\mathrm{ind}} - D}{\mathrm {MSE}_{\mathrm{ind}} + \mathrm {MSE}_{\mathrm{dir}} - 2D},
    \qquad 
\]
The H-learner strictly improves on both endpoints if and only if 
\( \omega^\star \in (0,1) \), which holds when
\begin{align}
    D < \min\{\mathrm {MSE}_{\mathrm{ind}},\,\mathrm {MSE}_{\mathrm{dir}}\} \label{eq:condition}
\end{align}
Now we prove Corollary 5.4 and 5.5:

Under the assumption of cross-fitting, $\Sigma_{\mathrm{ind,dir}} = \operatorname{Cov}(\hat{\theta}_{\mathrm{ind}}, \hat{\theta}_{\mathrm{dir}}) = 0, D = b_{\mathrm{ind}}^\top b_{\mathrm{dir}}$.

\begin{itemize}
    \item \textbf{Case 1}: If \(b_{\mathrm{ind}}^\top b_{\mathrm{dir}} < 0\), then \(D < 0 \le \min\{\mathrm{MSE}_{\mathrm{ind}}, \mathrm{MSE}_{\mathrm{dir}}\}\), 
    so condition \eqref{eq:condition} holds.

    \item \textbf{Case 2}: If $\|b_{\mathrm{ind}}\|_2 \ge \|b_{\mathrm{dir}}\|_2$ and $\operatorname{Var}(\hat{\theta}_{\mathrm{ind}}) = \operatorname{tr}(\Sigma_{\mathrm{dir}}) > b_{\mathrm{dir}}^\top ( b_{\mathrm{ind}} - b_{\mathrm{dir}} )$. Then
    \[
    D 
    = b_{\mathrm{ind}}^\top b_{\mathrm{dir}}
    \le \|b_{\mathrm{ind}}\|_2 \,\|b_{\mathrm{dir}}\|_2
    \le \|b_{\mathrm{ind}}\|_2^2
    < \|b_{\mathrm{ind}}\|_2^2 + \operatorname{tr}(\Sigma_{\mathrm{ind}})
    = \mathrm{MSE}_{\mathrm{ind}}
    \]
    \[
    D = b_{\mathrm{ind}}^\top b_{\mathrm{dir}}
    <
    b_{\mathrm{dir}}^\top b_{\mathrm{dir}} + \operatorname{tr}(\Sigma_{\mathrm{dir}})
    =
    \mathrm{MSE}_{\mathrm{dir}}
    \]
    Therefore \(D  < \min\{\mathrm{MSE}_{\mathrm{ind}}, \mathrm{MSE}_{\mathrm{dir}}\}\), 
    so condition \eqref{eq:condition} holds.
\end{itemize}

Since $\min_W \mathrm{MSE}_H(W)\le \min_{\omega\in[0,1]}\mathrm{MSE}_H(\omega I)$, the conditions derived above along this scalar path provide a sufficient guarantee: whenever the scalar path admits an interior minimizer, the full matrix optimization achieves no higher risk and therefore strictly improves upon both endpoints.

\section{Additional Semi-Synthetic Experiment Results}
\label{sec: add_syn}
\begin{figure*}[h]
  \centering
  \includegraphics[width=\textwidth]{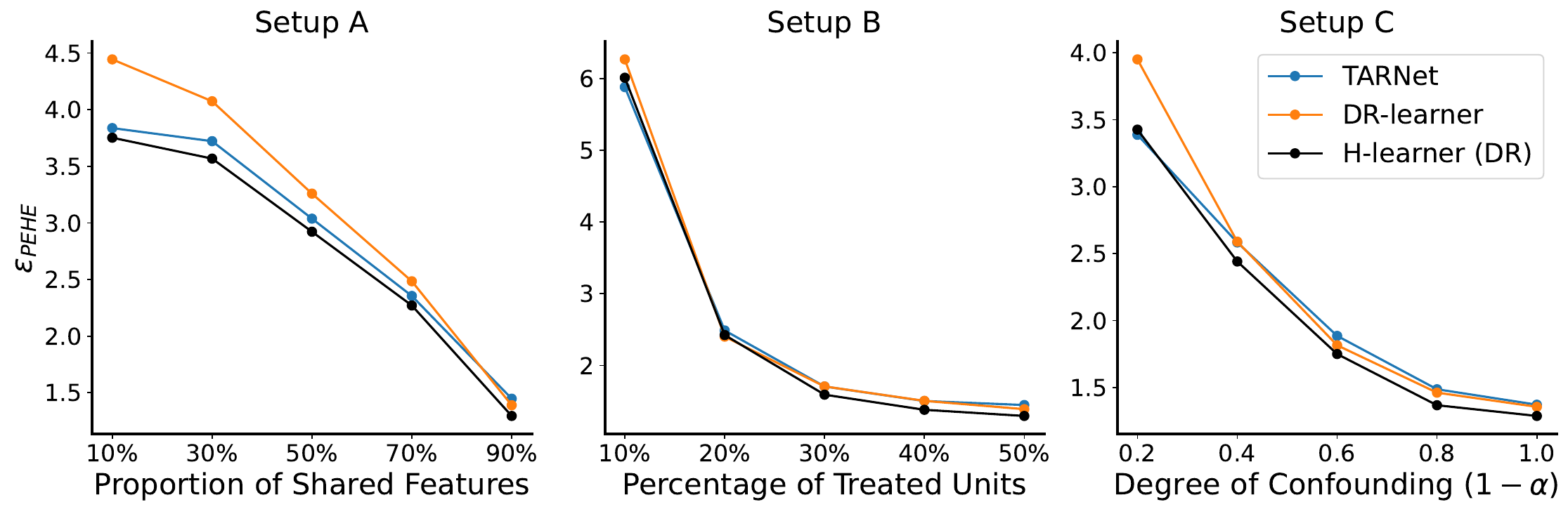}
    \caption{PEHE results for three semi-synthetic scenarios. The H-learner is more robust across these settings and outperforms both learners in most cases.}
  \label{fig:appx1}
\end{figure*}

In the main paper, we present results from the semi-synthetic experiment evaluating the H-learner (X). Here, we provide additional results for the H-learner (DR), comparing it against its counterparts: the DR-learner (direct learner) and TARNet (indirect learner). We follow the same experimental setup detailed in Section~\ref{subsec:synthetic} and fix the proportion of shared features at 90\% in Setups B and C to better observe the performance trade-offs. The results are summarized in Figure \ref{fig:appx1}.

\textbf{Results.} Setup A demonstrates the performance trade-off through the \textit{relative complexity of the POs and the CATE}. As the proportion of shared features increases, the CATE becomes simpler. Correspondingly, we observe that the DR-learner only outperforms TARNet when the proportion of shared features is high but underperform when the shared feature proportion is low. Setup B demonstrates the performance trade-off 
driven by the \textit{data imbalance between treated and control populations}. As the proportion of treated units decreases, the imbalance in sample sizes between the treatment and control groups becomes more severe. This affects the DR-learner, as its pseudo-outcomes depend on inverse propensity scores that exhibit high variance. Empirically, DR-learner performs better under balanced treatment assignment but underperforms when only 10\% of units receive treatment. Setup C demonstrates the trade-off associated with the \textit{strength of confounding}. As confounding increases and feature distributions diverge more significantly, the DR-learner performs worse due to exacerbated variance in its inverse propensity pseudo-outcomes. In contrast, the H-learner is more robust across these settings and performs better.

\begin{figure*}[h]
  \centering
  \includegraphics[width=\textwidth]{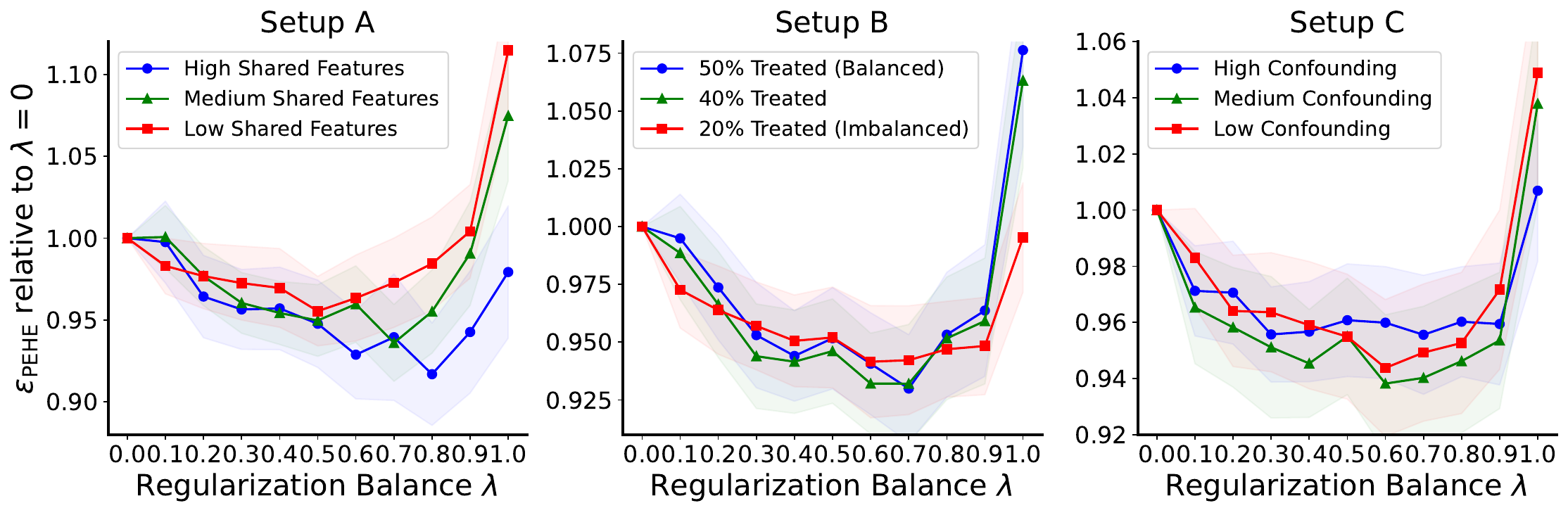}
    \caption{Performance of the H-learner across different $\lambda$ values for all three setups. Note that $\lambda = 0$ corresponds to the indirect learner (TARNet), and $\lambda = 1$ to the direct learner (X-learner). The H-learner consistently outperforms both learners across all three setups.}
  \label{fig:appx2}
\end{figure*}

\textbf{Effect of regularization parameter $\lambda$}. Lastly, we add additional results by expanding Figure~\ref{fig:synthetic2} to include all three experimental setups. Figure~\ref{fig:appx2} presents the H-learner’s performance across different values of the regularization strength $\lambda$ for each setup. Note that $\lambda = 0$ corresponds to the indirect learner, and $\lambda = 1$ to the direct learner. Depending on the levels of feature sharing, treatment proportion, and confounding, either $\lambda = 0$ or $\lambda = 1$ may perform better. However, the optimal value of $\lambda$ consistently falls between the two extremes, demonstrating the strength of the H-learner in leveraging both types of regularization. Across different setups, the H-learner consistently outperforms both the direct and indirect learners.

\section{Additional Results on ACIC 2016}
\label{sec:add_acic}
The ACIC 2016 benchmark datasets were introduced as part of the Atlantic Causal Inference Competition \citep{dorie2019automated} and offer a comprehensive testbed for evaluating causal inference methods. The benchmark includes 77 distinct DGPs that vary in the complexity of the response surface, the degree of confounding, covariate overlap, and treatment effect heterogeneity. While the main paper reports results averaged over all DGPs, here we provide aggregated results for specific settings.

\textbf{Results.} We present the results in Tables~\ref{tab:acic_2016_response}, 
\ref{tab:acic_2016_heterogeneity},
\ref{tab:acic_2016_confounding},
and~\ref{tab:acic_2016_overlap}. The H-learner consistently achieves state-of-the-art performance across various response models, levels of heterogeneity, confounding, and overlap. The best performance is obtained when constructing the X-learner pseudo-outcome in the first stage. We also observe that using the DR-learner pseudo-outcome yields competitive performance under settings with low confounding or low heterogeneity. This aligns with expectations, as the DR pseudo-outcome estimator tends to suffer less from high variance in such scenarios.

\begin{table}[H]
\centering
\small
\caption{Performance on the ACIC 2016 benchmark dataset across different \textbf{response models}.}
\vspace{1ex} 
\begin{tabular}{llcccccc}
\toprule
\multicolumn{2}{c}{Response Model} & \multicolumn{2}{c}{\textbf{Linear}} & \multicolumn{2}{c}{\textbf{Exponential}} & \multicolumn{2}{c}{\textbf{Step}} \\
\cmidrule(lr){3-4} \cmidrule(lr){5-6} \cmidrule(lr){7-8}
\multicolumn{2}{c}{} 
& \makecell{In-sample\\$\sqrt{\varepsilon_{\text{PEHE}}}$} 
& \makecell{Out-of-sample\\$\sqrt{\varepsilon_{\text{PEHE}}}$} 
& \makecell{In-sample\\$\sqrt{\varepsilon_{\text{PEHE}}}$} 
& \makecell{Out-of-sample\\$\sqrt{\varepsilon_{\text{PEHE}}}$}  
& \makecell{In-sample\\$\sqrt{\varepsilon_{\text{PEHE}}}$} 
& \makecell{Out-of-sample\\$\sqrt{\varepsilon_{\text{PEHE}}}$} \\
\midrule
& T-learner & $1.51 \pm .09$ & $1.56 \pm .11$ & $2.05\pm .05$ &  $2.19 \pm .05$ & $2.23 \pm .06$ & $2.30 \pm .06$  \\
& TARNet & $1.01 \pm .06$ & $1.03 \pm .07$ & $1.73 \pm .05$ & $1.86 \pm .05$ & $1.70 \pm .05$ & $1.79 \pm .06$ \\
\midrule
& IPW-learner & $1.83 \pm .17$ & $1.84 \pm .18$ & $3.13\pm .10$ &  $3.13 \pm .10$ & $2.31 \pm .09$ & $2.30 \pm .08$ \\
& DR-learner & $1.00 \pm .07$ & $1.03 \pm .09$ & $1.75 \pm .05$ & $1.90 \pm .06$ & $1.60 \pm .06$ & $1.69 \pm .06$ \\
& X-learner & $\bm{0.96 \pm .07}$ & $\bm{0.96 \pm .07}$ & $1.69\pm .05$ &  $1.84 \pm .05$ & $1.58 \pm .05$ & $1.67 \pm .06$ \\
\midrule
& TARNet-WR & $1.32 \pm .09$ & $1.35 \pm .11$ & $2.02\pm .06$ &  $2.12 \pm .06$ & $1.82 \pm .06$ & $1.89 \pm .06$ \\
& OffsetNet & $1.37 \pm .12$ & $1.38 \pm .12$ & $2.30 \pm .07$ & $2.37 \pm .07$ & $1.99 \pm .07$ & $2.03 \pm .08$ \\
& FlexTENet & $1.19 \pm .10$ & $1.23 \pm .12$ & $1.94 \pm .06$ & $2.04 \pm .06$ & $1.76 \pm .06$ & $1.83 \pm .07$ \\
\midrule
& \method\ (DR) & $\bm{0.96 \pm .07}$ & $1.00 \pm .09$ & $1.67 \pm .05$& $1.79 \pm .05$ & $1.57 \pm .06$ & $1.66 \pm .06$ \\
& \method\ (X) & $0.98 \pm .07$ & $1.00\pm .08$ &$\bm{1.65 \pm .04}$ & $\bm{1.78 \pm .05}$ & $\bm{1.56 \pm .06}$ & $\bm{1.65 \pm .06}$ \\
\bottomrule
\end{tabular}
\label{tab:acic_2016_response}
\end{table}

\begin{table}[H]
\centering
\small
\caption{Performance on the ACIC 2016 benchmark dataset across different \textbf{heterogeneity levels}.}
\vspace{1ex}
\begin{tabular}{llcccc}
\toprule
\multicolumn{2}{c}{Heterogeneity Level} & \multicolumn{2}{c}{\textbf{Low}} & \multicolumn{2}{c}{\textbf{High}} \\
\cmidrule(lr){3-4} \cmidrule(lr){5-6}
\multicolumn{2}{c}{} 
& \makecell{In-sample\\$\sqrt{\varepsilon_{\text{PEHE}}}$} 
& \makecell{Out-of-sample\\$\sqrt{\varepsilon_{\text{PEHE}}}$} 
& \makecell{In-sample\\$\sqrt{\varepsilon_{\text{PEHE}}}$} 
& \makecell{Out-of-sample\\$\sqrt{\varepsilon_{\text{PEHE}}}$}  \\
\midrule
& T-learner & $1.99 \pm .05$ & $2.06 \pm .06$ & $2.16\pm .05$ &  $2.30 \pm .06$\\
& TARNet & $1.56 \pm .05$ & $1.65 \pm .05$ & $1.77 \pm .05$ & $1.90 \pm .05$\\
\midrule
& IPW-learner & $2.47 \pm .11$ & $2.45 \pm .10$ & $2.95\pm .08$ &  $2.97 \pm .09$\\
& DR-learner & $1.51 \pm .05$ & $1.61 \pm .06$ & $1.77 \pm .05$ & $1.90 \pm .06$\\
& X-learner & $1.48 \pm .05$ & $1.57 \pm .05$ & $1.71\pm .05$ &  $1.85\pm .05$\\
\midrule
& TARNet-WR & $1.75 \pm .06$ & $1.82 \pm .06$ & $2.04\pm .06$ &  $2.13 \pm .06$\\
& OffsetNet & $1.91 \pm .07$ & $1.96 \pm .07$ & $2.31 \pm .06$ & $2.37 \pm .07$\\
& FlexTENet & $1.65 \pm .06$ & $1.72 \pm .06$ & $1.97 \pm .06$ & $2.08 \pm .06$\\
\midrule
& \method\ (DR) & $\bm{1.45 \pm .05}$ & $\bm{1.53 \pm .05}$ & $1.71 \pm .05$& $1.83 \pm .05$\\
& \method\ (X) & $\bm{1.45 \pm .05}$ & $\bm{1.53 \pm .05}$ &$\bm{1.69 \pm .05}$ & $\bm{1.82 \pm .05}$ \\
\bottomrule
\end{tabular}
\label{tab:acic_2016_heterogeneity}
\end{table}

\begin{table}[H]
\centering
\small
\caption{Performance on the ACIC 2016 benchmark dataset across different \textbf{confounding levels}.}
\vspace{1ex}
\begin{tabular}{llcccc}
\toprule
\multicolumn{2}{c}{Confounding Level} & \multicolumn{2}{c}{\textbf{Low}} & \multicolumn{2}{c}{\textbf{High}} \\
\cmidrule(lr){3-4} \cmidrule(lr){5-6}
\multicolumn{2}{c}{} 
& \makecell{In-sample\\$\sqrt{\varepsilon_{\text{PEHE}}}$} 
& \makecell{Out-of-sample\\$\sqrt{\varepsilon_{\text{PEHE}}}$} 
& \makecell{In-sample\\$\sqrt{\varepsilon_{\text{PEHE}}}$} 
& \makecell{Out-of-sample\\$\sqrt{\varepsilon_{\text{PEHE}}}$}  \\
\midrule
& T-learner & $2.09\pm .05$ &  $2.19 \pm .06$ & $2.08 \pm .05$ & $2.20 \pm .06$ \\
& TARNet & $1.64 \pm .05$ & $1.75 \pm .06$ & $1.69 \pm .05$ & $1.79 \pm .05$ \\
\midrule
& IPW-learner & $2.62\pm .10$ &  $2.59 \pm .10$ & $2.75 \pm .10$ & $2.76 \pm .10$ \\
& DR-learner & $1.61 \pm .05$ & $1.72 \pm .06$ & $1.66 \pm .05$ & $1.77 \pm .06$ \\
& X-learner & $1.57\pm .05$ &  $1.68\pm .06$ & $1.61 \pm .05$ & $1.72 \pm .06$ \\
\midrule
& TARNet-WR & $1.88 \pm .06$ &  $1.96 \pm .06$ & $1.91 \pm .06$ & $1.99 \pm .06$ \\
& OffsetNet & $2.05 \pm .07$ & $2.11 \pm .07$ & $2.15 \pm .07$ & $2.20 \pm .08$ \\
& FlexTENet & $1.79 \pm .06$ & $1.88 \pm .07$ & $1.83 \pm .06$ & $1.92 \pm .06$ \\
\midrule
& \method\ (DR) & $\bm{1.55 \pm .05}$& $\bm{1.65 \pm .06}$ & $1.60 \pm .05$ & $1.70 \pm .06$ \\
& \method\ (X) & $\bm{1.55 \pm .05}$ & $1.66 \pm .06$ & $\bm{1.58 \pm .05}$ & $\bm{1.68 \pm .05}$ \\
\bottomrule
\end{tabular}
\label{tab:acic_2016_confounding}
\end{table}

\begin{table}[H]
\centering
\small
\caption{Performance on the ACIC 2016 benchmark dataset across different \textbf{overlap levels}.}
\vspace{1ex}
\begin{tabular}{llcccc}
\toprule
\multicolumn{2}{c}{Overlap Level} & \multicolumn{2}{c}{\textbf{Full}} & \multicolumn{2}{c}{\textbf{Penalize}} \\
\cmidrule(lr){3-4} \cmidrule(lr){5-6}
\multicolumn{2}{c}{} 
& \makecell{In-sample\\$\sqrt{\varepsilon_{\text{PEHE}}}$} 
& \makecell{Out-of-sample\\$\sqrt{\varepsilon_{\text{PEHE}}}$} 
& \makecell{In-sample\\$\sqrt{\varepsilon_{\text{PEHE}}}$} 
& \makecell{Out-of-sample\\$\sqrt{\varepsilon_{\text{PEHE}}}$}  \\
\midrule
& T-learner & $2.09 \pm .05$ & $2.21 \pm .06$ & $2.07 \pm .05$ &  $2.17 \pm .05$\\
& TARNet & $1.66 \pm .05$ & $1.78 \pm .06$ & $1.66 \pm .05$ & $1.76 \pm .05$\\
\midrule
& IPW-learner & $2.73 \pm .10$ & $2.76 \pm .10$ & $2.65\pm .09$ &  $2.63 \pm .09$\\
& DR-learner & $1.63 \pm .05$ & $1.76 \pm .06$ & $1.64 \pm .05$ & $1.74 \pm .06$\\
& X-learner & $1.58 \pm .05$ & $1.71 \pm .06$ & $1.59\pm .05$ &  $1.69\pm .05$\\
\midrule
& TARNet-WR & $1.92 \pm .06$ & $2.01 \pm .07$ & $1.86 \pm .05$ &  $1.94 \pm .06$\\
& OffsetNet & $2.17 \pm .07$ & $2.23 \pm .08$ & $2.05 \pm .06$ & $2.09 \pm .07$\\
& FlexTENet & $1.85 \pm .06$ & $1.94 \pm .07$ & $1.78 \pm .05$ & $1.86 \pm .06$\\
\midrule
& \method\ (DR) & $1.57 \pm .05$ & $1.68 \pm .06$ & $1.58 \pm .05$& $\bm{1.67 \pm .05}$\\
& \method\ (X) & $\bm{1.55 \pm .05}$ & $\bm{1.66 \pm .06}$ &$\bm{1.57 \pm .05}$ & $\bm{1.67 \pm .06}$ \\
\bottomrule
\end{tabular}
\label{tab:acic_2016_overlap}
\end{table}

\end{document}